\documentclass{article} 
\usepackage{iclr2027_conference,times}

\usepackage{amsmath,amsfonts,bm}

\def\eqref#1{equation~\ref{#1}}

\def\1{\bm{1}}

\DeclareMathAlphabet{\mathsfit}{\encodingdefault}{\sfdefault}{m}{sl}
\SetMathAlphabet{\mathsfit}{bold}{\encodingdefault}{\sfdefault}{bx}{n}

\usepackage{adjustbox}
\usepackage{booktabs}
\usepackage{multirow}
\usepackage{graphicx}
\usepackage{wrapfig}
\usepackage{subcaption}
\usepackage{amsmath,amssymb}
\usepackage{enumitem}
\usepackage{xcolor}
\usepackage{colortbl}
\usepackage{wasysym}
\usepackage{hyperref}
\usepackage{listings}

\lstdefinelanguage{json}{
    basicstyle=\ttfamily\footnotesize,
    showstringspaces=false,
    breaklines=true,
    breakautoindent=false,
    breakindent=0pt,
    columns=fullflexible,
    keepspaces=true,
    showstringspaces=false,
    framexleftmargin=4pt,
    framexrightmargin=4pt,
    frame=lines,
    backgroundcolor=\color{gray!10},
    morestring=[b]`,
    stringstyle=\color{brown},
    literate=
     *{0}{{{\color{blue}0}}}{1}
      {1}{{{\color{blue}1}}}{1}
      {2}{{{\color{blue}2}}}{1}
      {3}{{{\color{blue}3}}}{1}
      {4}{{{\color{blue}4}}}{1}
      {5}{{{\color{blue}5}}}{1}
      {6}{{{\color{blue}6}}}{1}
      {7}{{{\color{blue}7}}}{1}
      {8}{{{\color{blue}8}}}{1}
      {9}{{{\color{blue}9}}}{1}
}

\newcommand{\fullmark}{\ensuremath{\CIRCLE}}
\newcommand{\halfmark}{\ensuremath{\LEFTcircle}}
\newcommand{\emptymark}{\ensuremath{\Circle}}
\newcommand{\unknownmark}{\textbf{?}}
\newcommand{\bestf}[1]{\cellcolor{red!15}\textbf{#1}}
\newcommand{\besta}[1]{\cellcolor{blue!15}\textbf{#1}}

\title{Using LLMs to Detect LLM-Generated Texts: A Cross-Generation Analysis}

\author{%
  Haiyue Yuan$^1$,
  Jie Guo$^2$,
  Weidong Qiu$^2$,
  \textbf{Zheng Huang}$^2$,
  \textbf{Ruizhe Li}$^3$\thanks{\ Corresponding co-authors: \texttt{r.li.7@bham.ac.uk, s.j.li@kent.ac.uk}} ,
  \textbf{Shujun Li}$^1$\footnotemark[1]\\
  $^1$Institute of Cyber Security for Society (iCSS) \& School of Computing, University of Kent, UK\\
  $^2$School of Cyber Science and Engineering, Shanghai Jiao Tong University, China\\
  $^3$School of Computer Science, University of Birmingham, UK\\
}

\iclrfinalcopy 
\begin{document}

\maketitle

\begin{abstract}
Automated detection of LLM-generated texts (LGTs) is critical, yet dedicated detectors often struggle to generalize across domains and models. While general-purpose LLMs offer flexible zero-shot authorship classification with explanatory rationale, their detection behavior, especially regarding self-detection versus cross-detection across model generations, remains poorly understood. We systematically evaluate 15 LLMs spanning three model generations as both generators and detectors. Using a benchmark of 1,000 human-written texts and 15,000 LGTs (1,000 per model), we collected over 233,000 binary classifications alongside natural-language explanations. Our results reveal that detection efficacy is primarily driven by detector capability rather than generator provenance, although outputs from newer generators remain notably harder to detect. Crucially, statistical comparisons show no systematic advantage or disadvantage for self-detection across models. Error analysis further exposes generational bias shifts: first-generation detectors under-detect LGTs (high false-negative rates), second-generation detectors over-flag human texts (high false-positive rates), and the latest models achieve balanced trade-offs. Finally, we highlight significant inconsistencies in how different LLMs apply textual cues to justify their decisions.
\end{abstract}

\section{Introduction}
\label{sec:introduction}

In recent years, large language models (LLMs) have developed from research prototypes into various applications used across a wide range of domains, such as education, science, medicine, programming, journalism, and creative work~\citep{Li-J2024}. While these applications can help improving productivity and work efficiency, the same generative capability enables LLMs to produce convincing content that could reinforce false narratives, raising concerns about the generation and dissemination of misinformation and disinformation at scale~\citep{Vykopal-I2024}. It may also be misused to create personalised phishing messages and deceptive online reviews~\citep{Czybik-S2026, Kovacs-B2024}. On the other hand, falsely accusing legitimate authors of using LLMs can also cause reputational harm~\citep{Liang-W2023, Layton-S2026}. 

LLM-generated text (LGT) detection has therefore emerged as an important research topic, answering questions of textual provenance, plagiarism, and distinction between human- and machine- generated content~\citep{Dugan-L2024, Lee-J2023, Bhattacharjee-A2024, Wu-J2025b}. Prior work spans supervised classifiers, statistical methods, rewriting-based approaches, and watermarking~\citep{Mitchell-E2023, Hans-A2024, Kirchenbauer-J2023}. Various efforts have also been made to evaluate large benchmarks and commercial detectors for detecting LGTs~\citep{Wang-Y2024, Dugan-L2024, Layton-S2026}. Despite such substantial efforts from both research and industry communities, such detectors often generalise poorly across unseen models and domains, while they remain vulnerable to prompting and style-shifting attacks~\citep{Tufts-B2025, Pedrotti-A2025}. It is also worth noting that when detectors are required to keep low false positive rate, the performance of detecting LGT decline significantly~\citep{Tufts-B2025}. Moreover, detection becomes even more challenging for human-LLM co-authored texts, where detectors need to explicitly identify the source of individual segments~\citep{Su-Z2025}.

An increasingly attractive approach is to simply use a general-purpose LLM as the detector, where the selected LLM takes a suspected text and returns a binary decision~\citep{Hans-A2024, Ji-J2025}. It can also provide a natural-language explanation on top of the binary decision label~\citep{Ji-J2025}. The study closest to ours was conducted by \citet{Ji-J2025}, where six open- and closed-source LLMs were evaluated as binary and ternary detectors. Their study suggested that an LLM generally performs better at identifying texts generated by itself compared with recognising texts generated by other LLMs. Slightly differently, \citet{Leinonen-J2026} examined self-detection by a single LLM, and found that self-detection varies substantially across task, response length, and prompt formulation. Both studies have limited scopes and their results need more independent validation with more recent LLMs, to enhance our understanding on self- and cross-detection of LGTs. More specifically, neither studied generational effects, e.g., how LLMs of different generations behave as self- and cross-detector of LGTs.

To fill the above-mentioned research gaps, we conducted a large-scale study of LGT detection using 15 general-purpose LLMs of three generations (i.e., \emph{G1: 2023-24}, \emph{G2: 2024-25}, and \emph{G3: 2026}) as both text generators and detectors. Following~\citet{Ji-J2025}'s approach, we selected 1,000 human-generated texts (HGTs) from the publicly available M4GT-Bench dataset~\citep{Wang-Y2024}. Next, each of the 15 selected LLMs was used to generate 1,000 LGTs. The task is to use each selected LLM as a detector to perform binary classification task on the 15,000 LGTs and 1,000 HGTs and to provide a natural-language explanation for each of its decisions. Our work led to the following main findings and contributions:
\begin{enumerate}[itemindent=0pt, leftmargin=*]
\item \textbf{A controlled and openly accessible benchmark.} We constructed a corpus containing 1,000 HGTs and 15,000 LGTs produced by 15 general-purpose LLMs. Applying every model as a detector results in a complete matrix of 225 generator--detector combinations and 233,428 valid binary judgements with natural-language explanations.

\item \textbf{A cross-generational evaluation of LLM detectors.} We defined and introduced three LLM generations (based on recency) and separately examined how generator and detector generations affect detection performance. Detection performance improves substantially over time, and detector generation is more strongly associated with performance than generator generation, although texts generated G3 generators are generally harder to detect, reflecting the more improved capabilities of more recent LLMs.

\item \textbf{A more comprehensive study of self- and cross-detection.} Across all generator and detector generations, there are only slight differences in average performances between self- and cross-detection, while the direction of the difference varies across different detectors. We also conducted a controlled comparison between each detector's self-detection and cross-detection performance from the same generation, using prompt-clustered bootstrap tests, Holm correction, and sensitivity analyses. We have observed both significant self-detection advantages and disadvantages for different detectors within the same generation, however, their effects are generally small and do not support a consistent and conclusive self-detection advantage or disadvantage across LLMs. These results challenge the results shown in~\citep{Ji-J2025}.

\item \textbf{A multi-perspective analysis of error patterns and explanations.} We analysed error patterns from three perspectives: false-positive rate (FPR) and false-negative rate (FNR) analysis, an LLM-assisted analysis of explanations, and a literature-guided cue-based analysis. The first analysis revealed distinct generational patterns: G1 detectors show a strong ``everything is human-written'' tendency, G2 detectors more frequently predict machine, and most G3 detectors maintain a low FPR and a low FNR. The two subsequent analyses of semantic error patterns converge in showing that \emph{structured and generic writing} is commonly associated with LGTs, whereas \emph{specificity} can lead to a LGT being misclassified as a HGT, and \emph{coherence} is associated with both HGT and LGT judgements.
\end{enumerate}

\section{Related Work}
\label{sec:related-work}

\textbf{LLM-generated text detection}: Prior work developed supervised, training-free, rewriting-based, and watermarking approaches to LGT detection as a binary classification task~\citep{Wu-J2025a, Mitchell-E2023, Hans-A2024, Huang-Y2025}. More recent methods also use linguistic, semantic, and structural features~\cite{Wu-J2025b, Guo-R2026, Xiong-R2026, Wu-C2026}. Although these methods can perform well under matched conditions, their reliability often decreases across unfamiliar generators, domains, decoding settings, and prompt styles~\citep{Wang-Y2024, Dugan-L2024, Tufts-B2025, Layton-S2026}.

\textbf{LLMs as detectors and self-detection}: General-purpose LLMs can also be used as zero-shot detectors that detect LGT and HGT with a natural-language explanation. However, their predictions may be strongly biased towards one class~\citep{Bhattacharjee-A2024}. \citet{Ji-J2025} reported a self-detection advantage across six LLMs, but subsequent evidence from other research produced mixed results, with self-recognition varying across models, evaluation settings, task domains, prompts, response lengths, and writing styles~\citep{Panickssery-A2024, Bai-X2025, Caiado-A2025, StAmand-J2026, Leinonen-J2026}.

\textbf{Error analysis and explanations}: Natural-language explanations can reveal the cues used by LLM detectors, but a correct label may still be supported by weak, absent, or incorrectly interpreted evidence~\citep{Ji-J2025}. Human annotators similarly rely on cues such as vocabulary, structure, grammar, and originality, although the same cue may support different judgement~\citep{Russell-J2025}. Research on commercial detector can provide only limited insight because many reported cues overlap between HGTs and LGTs~\citep{Ji-J2024}.

\textbf{Research gaps} Systematic evaluations of general-purpose LLMs acting as both generators and detectors remain limited in scale, leaving self- and cross-detection across broader model generations insufficiently understood~\citep{Ji-J2025}. It also remains unclear how the cues reported by LLM detectors relate to correct decisions, false positives, and false negatives across these settings. A more detailed review of the individual methods and findings is provided in Appendix~\ref{subsec:appendix-related-work}.

\section{Experimental design}
\label{sec:experimental-design}

\textbf{Model generations and selections.} 
We grouped models into three generations largely based on time of release: G1 (2023 and early 2024), G2 (2024-25), G3 (2026). G1 covers early-stage LLMs designed primarily to follow input prompts and output direct responses. G2 covers more powerful general-purpose LLMs with improved instruction following, coding, multilingual, long-context, and general reasoning capabilities, but with minor inference-time thinking or reasoning. G3 cover the most recent models (at the time of our experiments) that are designed for inference-time reasoning with extended thinking, tool use, and agentic execution capabilities.

The model selection for each generation follows the same rule. They must support general-purpose text generation and judgement, provide diversity across vendors, and collectively include both proprietary and open-weight models. It is worth noting that we mainly accessed the models via OpenRouter and vendors' native APIs. Since the fast development of LLMs in general, many early candidate models have been deprecated or discontinued. Hence, model availability constrains the final model selection, resulting in fewer G1 models and unbalanced group sizes across generations. The final selection is summarised in Table~\ref{tab:model-generation-selection}.

\begin{table}[!tbh]
\centering
\caption{Selected models across generations. Every model acts as both a text generator and a detector. Models with multimodal interfaces are evaluated using text only.}
\label{tab:model-generation-selection}
\scriptsize
\begin{tabular}{c p{0.85\linewidth}}
\toprule
Generation & Selected models\\
\midrule
G1 & GPT-3.5 Turbo Instruct~\citep{OpenAI-O2023}; Mixtral 8$\times$22B Instruct~\citep{Mistral-M2024a}.\\
\addlinespace
G2 & GPT-4o~\citep{OpenAI-O2024}; Qwen2.5-72B Instruct~\citep{Qwen-Q2024}; DeepSeek-V3~\citep{DeepSeek-D2024};
Llama 3.3 70B Instruct~\citep{Meta-M2024}; GLM-4 32B 0414~\citep{ZAI-Z2025}.\\
\addlinespace
G3 &  DeepSeek-V4-Pro~\citep{DeepSeek-D2026}; Qwen3.6-Max-Preview~\citep{Qwen-Q2026}; GLM-5.1~\citep{ZAI-Z2026}; MiMo-V2.5-Pro~\citep{Xiaomi-X2026}; GPT-5.5~\citep{OpenAI-O2026}; Claude Opus 4.7~\citep{Anthropic-A2026}; Gemini 3.1 Pro Preview~\citep{Google-G2026};
Kimi K2.6~\citep{Moonshot-M2026}.\\
\bottomrule
\end{tabular}
\end{table}

\textbf{Corpus construction.} 
Following~\citep{Ji-J2025}, we directly reused their HGT dataset and the corresponding text generation prompts reported in their study to populate the LGT dataset. Specifically, the HGT dataset contains 1,000 HGTs selected from the publicly available M4GT-Bench dataset~\citep{Wang-Y2024}, covering a broad range of topics, styles, and formats. Based on these HGTs, 1,000 matching text-generation prompts were created to align with the themes, structures, and styles appearing in the 1,000 HGTs. See Listing~\ref{lst:generation-prompt-example} in Appendix~\ref{subsubsec:appendix-curpus-construction} for an example prompt and refer to \citep{Ji-J2025} for more details. Then, we applied the 1,000 text-generation prompts to each of the 15 selected LLMs, with each model generating one LGT for each prompt. See Appendix~\ref{subsubsec:appendix-curpus-construction}for a example LGT generated by GPT-5.5 for an example prompt.

For text generation, models were accessed through \citet{OpenRouter-O2026} using an OpenAI-compatible chat-completions interface. Each request contains the matched user prompt and a fixed system instruction requiring plain, continuous prose without headings, lists, or Markdown formatting. The user prompts (i.e., text-generation prompts) specify the topic, task, expected form, and a target length, which typically contains 150--300 words. In addition, we reviewed the documented default temperature and top-$p$ values for all selected models, as reported in Table~\ref{tab:llm_parameters} in Appendix~\ref{subsubsec:appendix-curpus-construction}. A temperature of 1.0 and top-$p$ of 1.0 or close to 1.0, such as 0.95, are the most commonly used values. Hence, we set both temperature and top-$p$ to 1.0 as common experimental settings to ensure comparable generation conditions across models. In total, with 15 selected LLM models as text generators, we produced a corpus that contains of 1,000 HGTs and 15,000 (i.e., $15\,\text{generators} \times 1,000\,\text{prompts}$) LGTs, comprising 16,000 texts in total. The corpus can be accessed from~\url{https://github.com/hyyuan/detect-llm-generated-texts}.

\textbf{Binary classification on LGTs and HGTs.} 
After constructing the corpus, we used the selected LLMs as detectors to perform a binary classification task to judge whether a given text was written by a human or generated by an LLM. For each generator, we combined its 1,000 generated LGTs with the same set of 1,000 HGTs to construct a generator-specific dataset. Hence, we have 15 generator-specific datasets, each containing 2,000 texts. Each detector then performs the binary classification task on all 15 generator-specific datasets. In total, this process generates 240,000 binary classification results (i.e., 15 detectors $\times$ 15 datasets $\times$ 1000 LGTs $+$ 15 detectors $\times$ 1000 HGTs). When the detector and generator are the same model, the evaluation is considered self-detection; otherwise, it is considered cross-detection.

For the binary classification task, the same zero-shot instruction was used for every detector, where the instruction explicitly asks the detector to provide both a binary judgement with a natural-language explanations to support its decision. For each request made by each detector, the text to be judged is appended to this instruction, with no additional system prompt. The required JSON response only contains two fields. The \texttt{answer} field must contain either \texttt{human} or \texttt{machine}, while the \texttt{explanation} field records the corresponding detector's explanations in natural language. See Appendix~\ref{subsubsec:appendix-binary-classification} for an example detection result generated by GPT-5.5.

\textbf{Evaluation metrics.}
Treating LGT as the positive class, F1-score and accuracy are the primary metrics used to evaluate classification performance. FPR and FNR are also used to further interpret the results and identify class-specific error patterns. Of the 240,000 classification responses, 233,428 (97.3\%) produced valid binary judgements after excluding 1,909 missing responses and 4,663 malformed responses (more details in Appendix~\ref{subsubsec:appendix-selfcross-statistics}).

\section{Results}
\label{sec:results}

\subsection{Overall performance}
\label{subsec:overall-performance}

Table~\ref{tab:full-results} shows detection performance in a 15$\times$15 generator--detector matrix. Each generator represents its corresponding generator-specific dataset and has one row for F1-score and one row for accuracy. Red and blue values denote F1-score and accuracy for self-detection, respectively. For each detector column, the highest F1-score is shown in bold with a light-red background, while the highest accuracy is shown in bold with a light-blue background. Tied maximum values are all highlighted. Horizontal rules separate G1, G2, and G3 generators, while vertical rules separate the three detector generations. For simplicity and readability, we use DS to denote DeepSeek, Claude to denote Claude Opus 4.7, and MiMo to denote MiMo V2.5 Pro in the table. Across the matrix, the highest column-wise values are primarily achieved by G3 detectors, particularly on texts generated by G1 and G2 models as illustrated in the table. The mean F1-score is 69.5\% and the mean accuracy is 76.5\%. However, both metrics vary substantially, with standard deviations of 31.7 and 19.6 percentage points, respectively. To this end, the overall averages of F1-score and accuracy indicate the considerable variation across model pairings rather than overall strong detection performance.

\begin{table}[htbp!]
\centering
\caption{Complete classification results for all generator--detector pairings (\%).}
\label{tab:full-results}
\setlength{\tabcolsep}{1.8pt}
\resizebox{\linewidth}{!}{%
\begin{tabular}{|*{18}{c|}}
\hline
Generation & Generator & Metric & \rotatebox{60}{GPT-3.5} & \rotatebox{60}{Mixtral} & \rotatebox{60}{DS-V3} & \rotatebox{60}{GLM-4} & \rotatebox{60}{GPT-4o} & \rotatebox{60}{Llama-3.3} & \rotatebox{60}{Qwen2} & \rotatebox{60}{Claude} & \rotatebox{60}{DS-V4} & \rotatebox{60}{Gemini-3.1} & \rotatebox{60}{GLM-5.1} & \rotatebox{60}{GPT-5.5} & \rotatebox{60}{Kimi-K2.6} & \rotatebox{60}{MiMo-V2.5} & \rotatebox{60}{Qwen3.6}\\
\hline
\multirow{4}{*}{G1} & \multirow{2}{*}{GPT-3.5} & F1 & \textcolor{red}{3.0} & 42.6 & 62.7 & 59.1 & 80.6 & \bestf{77.1} & 1.7 & 97.0 & 93.1 & \bestf{99.9} & 84.7 & \bestf{99.0} & 99.3 & 91.9 & 99.0\\
& & Accuracy & \textcolor{blue}{47.4} & 56.5 & 60.6 & 61.2 & 76.6 & \besta{71.1} & 48.3 & 96.9 & 93.0 & \besta{99.9} & 86.5 & \besta{99.0} & 99.3 & 91.9 & 99.0\\
\cline{2-18}
& \multirow{2}{*}{Mixtral} & F1 & 2.3 & \textcolor{red}{38.6} & 62.0 & 51.3 & 75.4 & 70.0 & 0.6 & 96.8 & 82.9 & 99.7 & 80.2 & 98.9 & 99.2 & 86.1 & 99.1\\
& & Accuracy & 47.3 & \textcolor{blue}{54.7} & 60.1 & 56.2 & 71.6 & 64.2 & 47.9 & 96.7 & 84.1 & 99.7 & 83.2 & 98.9 & 99.2 & 86.8 & 99.1\\
\hline
\multirow{10}{*}{G2} & \multirow{2}{*}{DS-V3} & F1 & 2.4 & 39.5 & \textcolor{red}{51.9} & 53.8 & 76.5 & 73.0 & 1 & 97.0 & 88.1 & 99.8 & 84.5 & 99.0 & 99.2 & 88.3 & \bestf{99.2}\\
& & Accuracy & 47.4 & 55.0 & \textcolor{blue}{52.9} & 57.8 & 72.6 & 67.0 & 48.0 & 96.9 & 88.4 & 99.8 & 86.4 & 98.9 & 99.2 & 88.7 & \besta{99.2}\\
\cline{2-18}
& \multirow{2}{*}{GLM-4} & F1 & 3.4 & 38.0 & 56.1 & \textcolor{red}{53.9} & 75.6 & 71.8 & 0.6 & 96.9 & 85.3 & 99.7 & 83.1 & 98.9 & 99.0 & 84.9 & 98.9\\
& & Accuracy & 47.5 & 54.4 & 55.7 & \textcolor{blue}{57.9} & 71.8 & 65.8 & 47.9 & 96.8 & 86.1 & 99.7 & 85.3 & 98.9 & 99.0 & 85.9 & 98.9\\
\cline{2-18}
& \multirow{2}{*}{GPT-4o} & F1 & 1.7 & 40.6 & 60.7 & 55.3 & \textcolor{red}{77.3} & 73.4 & 0.4 & 97.0 & 89.7 & 99.8 & \bestf{89.1} & 98.9 & 99.2 & 89.9 & 99.2\\
& & Accuracy & 47.0 & 55.6 & 59.1 & 58.7 & \textcolor{blue}{73.4} & 67.4 & 47.9 & 96.9 & 89.9 & 99.8 & \besta{90.1} & 98.9 & 99.2 & 90.1 & 99.2\\
\cline{2-18}
& \multirow{2}{*}{Llama-3.3} & F1 & 3.0 & 47.9 & \bestf{71.1} & 60.5 & \bestf{80.8} & \textcolor{red}{76.8} & 4.7 & 97.0 & \bestf{93.4} & 99.8 & 85.2 & 99.0 & 99.4 & \bestf{92.1} & \bestf{99.2}\\
& & Accuracy & 47.4 & 59.2 & \besta{67.5} & 62.2 & \besta{76.8} & \textcolor{blue}{70.8} & 49.0 & 96.9 & \besta{93.2} & 99.9 & 86.9 & 98.9 & 99.4 & \besta{92.2} & \besta{99.2}\\
\cline{2-18}
& \multirow{2}{*}{Qwen2} & F1 & 4.8 & 44.1 & 64.5 & 58.9 & 80.3 & 76.9 & \textcolor{red}{1.7} & \bestf{97.0} & 91.5 & \bestf{99.9} & 87.2 & 99.0 & 99.4 & 92.0 & 99.1\\
& & Accuracy & 47.9 & 57.3 & 62.0 & 61.1 & 76.3 & 70.9 & \textcolor{blue}{48.2} & \besta{96.9} & 91.5 & \besta{99.9} & 88.5 & 98.9 & 99.4 & 92.0 & 99.1\\
\hline
\multirow{16}{*}{G3} & \multirow{2}{*}{Claude} & F1 & 2.6 & 31.6 & 31.4 & 44.4 & 66.6 & 57.7 & 0.6 & \textcolor{red}{96.7} & 77.6 & 98.9 & 78.0 & 98.8 & 98.6 & 79.8 & 97.7\\
& & Accuracy & 47.4 & 51.6 & 40.9 & 52.3 & 63.9 & 53.8 & 47.9 & \textcolor{blue}{96.6} & 80.2 & 98.9 & 81.7 & 98.8 & 98.6 & 81.9 & 97.8\\
\cline{2-18}
& \multirow{2}{*}{DS-V4} & F1 & 5.0 & 38.8 & 57.2 & 46.7 & 71.6 & 60.3 & 1.1 & 95.7 & \textcolor{red}{76.2} & 99.3 & 77.8 & 98.7 & 98.8 & 79.3 & 97.2\\
& & Accuracy & 48.0 & 54.8 & 56.5 & 53.6 & 68.2 & 55.9 & 48.1 & 95.6 & \textcolor{blue}{79.1} & 99.3 & 81.6 & 98.7 & 98.8 & 81.5 & 97.3\\
\cline{2-18}
& \multirow{2}{*}{Gemini-3.1} & F1 & 9.0 & \bestf{63.4} & 61.1 & 52.0 & 73.3 & 73.3 & 3.9 & 95.7 & 76.2 & \textcolor{red}{93.9} & 84.4 & 98.7 & 97.3 & 66.3 & 94.2 \\
& & Accuracy & 49.1 & \besta{68.1} & 59.4 & 56.7 & 69.7 & 67.3 & 48.9 & 95.6 & 79.2 & \textcolor{blue}{94.2} & 86.3 & 98.8 & 97.3 & 72.8 & 94.5\\
\cline{2-18}
& \multirow{2}{*}{GLM-5.1} & F1 & 11.3 & 59.0 & 51.5 & \bestf{69.0} & 79.7 & 76.0 & \bestf{14.9} & 96.2 & 70.4 & 98.6 & \textcolor{red}{72.7} & 98.3 & 98.5 & 73.8 & 97.2\\
& & Accuracy & 49.9 & 65.3 & 52.6 & \besta{68.4} & 75.7 & 69.9 & \besta{52.0} & 96.1 & 75.2 & 98.6 & \textcolor{blue}{78.2} & 98.4 & 98.5 & 77.6 & 97.3\\
\cline{2-18}
& \multirow{2}{*}{GPT-5.5} & F1 & 1.7 & 31.4 & 58.9 & 50.0 & 72.4 & 66.6 & 0.4 & 96.9 & 76.9 & 99.7 & 82.0 & \textcolor{red}{98.9} & 98.5 & 82.3 & 98.2\\
& & Accuracy & 47.1 & 51.5 & 57.7 & 55.5 & 68.8 & 61.2 & 47.9 & 96.8 & 79.6 & 99.7 & 84.6 & \textcolor{blue}{98.8} & 98.5 & 83.8 & 98.3\\
\cline{2-18}
& \multirow{2}{*}{Kimi-K2.6} & F1 & \bestf{15.2} & 39.9 & 53.9 & 49.9 & 78.9 & 58.8 & 4.9 & 92.4 & 68.1 & 99.4 & 85.5 & 98.9 & \textcolor{red}{98.9} & 71.6 & 96.2\\
& & Accuracy & \besta{50.9} & 55.3 & 54.2 & 55.3 & 74.9 & 54.7 & 49.1 & 92.5 & 73.9 & 99.4 & 87.2 & 98.9 & \textcolor{blue}{98.9} & 76.1 & 96.3\\
\cline{2-18}
& \multirow{2}{*}{MiMo-V2.5} & F1 & 3.4 & 36.1 & 34.0 & 47.3 & 69.3 & 65.0 & 0.8 & 96.8 & 81.1 & 99.2 & 82.6 & 98.9 & 98.9 & \textcolor{red}{84.6} & 98.7\\
& & Accuracy & 47.5 & 53.5 & 42.3 & 53.9 & 66.1 & 59.7 & 48.0 & 96.7 & 82.7 & 99.2 & 85.0 & 98.9 & 98.9 & \textcolor{blue}{85.6} & 98.7\\
\cline{2-18}
& \multirow{2}{*}{Qwen3.6} & F1 & 3.9 & 36.4 & 64.3 & 50.5 & 73.9 & 67.5 & 0.2 & 96.9 & 80.7 & 99.6 & 77.3 & 98.9 & \bestf{99.4} & 83.0 & \textcolor{red}{99.0}\\
& & Accuracy & 47.7 & 53.6 & 61.8 & 55.8 & 70.2 & 62.0 & 47.9 & 96.8 & 82.6 & 99.6 & 81.3 & 98.9 & \besta{99.4} & 84.3 & \textcolor{blue}{99.0}\\
\hline
\end{tabular}%
}
\end{table}

To further examine the variation, we also looked at the mean performance of each detector across 15 generator-specific dataset (see Table~\ref{tab:mean-performance-detector-generator} Panel A in Appendix~\ref{subsubsec:appendix-overall-performance}), as well as the mean performance of all detectors on each generator-specific dataset (see Table~\ref{tab:mean-performance-detector-generator} Panel B in Appendix~\ref{subsubsec:appendix-overall-performance}). Gemini 3.1 Pro Preview exhibits the highest performance across all generator-specific datasets, with a mean F1-score of 99.2\% and mean accuracy of 99.2\%, whereas Qwen2-72B has a mean F1-score of 2.5\% and mean accuracy of 48.5\%. In addition, across all detectors, texts generated by Llama 3.3 70B are the easiest to classify across all detectors, with a mean F1-score of 74\% and mean accuracy of 80\%, whereas texts generated by Claude Opus 4.7 are the most difficult to classify, with a mean F1-score of 64.1\% and mean accuracy of 72.8\%.

\subsection{Cross-generation analysis}
\label{subsec:cross-generation-analysis}

As shown in Table~\ref{tab:generation-performance}, detection performance improves substantially from G1 to G3 detectors across all generator-specific datasets. On average, the F1-score improves from 22.8\% to 93.1\% and the accuracy increases from 52.0\% to 93.7\% from G1 to G3. In particular, G1 detectors achieved low F1-score despite the accuracy slightly over 50.0\%, suggesting that G1 detectors frequently failed to identify LGTs rather than making balanced predictions across two classes. 

\begin{table}[htb!]
\centering
\caption{Detection performance across model generations (\%).}
\begin{subtable}[t]{0.4\linewidth}
\centering
\caption{Mean classification performance for each pairing of generator and detector generations}
\label{tab:generation-performance}
\adjustbox{max width=\linewidth}{
\begin{tabular}{ccccc}
\toprule
& & \multicolumn{3}{c}{Detector Generation}\\
\cmidrule(lr){3-5}
Generator Generation & Metric & G1 & G2 & G3\\
\midrule
\multirow{2}{*}{G1} & F1 & 21.6 & 54.1 & 94.2\\
 & Accuracy & 51.5 & 61.8 & 94.6\\
\addlinespace
\multirow{2}{*}{G2} & F1 & 22.5 & 54.3 & 94.9\\
 & Accuracy & 51.9 & 61.9 & 95.1\\
\addlinespace
\multirow{2}{*}{G3} & F1 & 24.3 & 49.0 & 90.3\\
 & Accuracy & 52.6 & 57.7 & 91.5\\
\midrule
\multirow{2}{*}{Macro average} & F1 & 22.8 & 52.5 & 93.1\\
 & Accuracy & 52.0 & 60.5 & 93.7\\
\bottomrule
\end{tabular}
}
\end{subtable}
\hfill
\begin{subtable}[t]{0.55\linewidth}
\centering
\caption{Self- and cross-detection performance across model generations}
\label{tab:selfcross}
\adjustbox{max width=\linewidth}{
\begin{tabular}{cccccc}
\toprule
Setting & Generator Generation & Metric & \multicolumn{3}{c}{Detector and Generator Generation} \\
\cmidrule(lr){4-6}
& & & G1 & G2 & G3\\
\midrule
\multirow{2}{*}{Self} & \multirow{2}{*}{Same} & F1 & 20.8 & 52.3 & 90.1\\
 & & Accuracy & 51.1 & 60.6 & 91.3\\
\midrule
\multicolumn{3}{c}{Cross-Detection by Generator Generation} & \multicolumn{3}{c}{Detector Generation} \\
\cmidrule(lr){4-6}
& & & G1 & G2 & G3\\
\midrule
\multirow{6}{*}{Cross} & \multirow{2}{*}{G1} & F1 & 22.4 & 54.1 & 94.2\\
 & & Accuracy & 51.9 & 61.8 & 94.6\\
\addlinespace
 & \multirow{2}{*}{G2} & F1 & 22.5 & 54.8 & 94.9\\
 & & Accuracy & 51.9 & 62.3 & 95.1\\
\addlinespace
 & \multirow{2}{*}{G3} & F1 & 24.3 & 49.0 & 90.4\\
 & & Accuracy & 52.6 & 57.7 & 91.5\\
\bottomrule
\end{tabular}
}
\end{subtable}
\end{table}

When comparing generations of generators, we observed a more subtle but still visible pattern. G2 detectors perform similarly on texts generated by G1 and G2 models, but their F1-scores decrease from approximately 54\% to 49\% and their accuracy values decrease from approximately 62\% to 58\% on G3-generated texts. G3 detectors show a similar pattern, achieving better performance on texts generated by G1 and G2 models than on those generated by G3 models. In contrast, G1 detectors consistently perform poorly on texts produced by all three generations of generators. These results suggest that detection performance is more strongly associated with detector generation than with generator generation. Nevertheless, texts produced by G3 models appear more difficult to classify, even for the more capable G2 and G3 detectors.

\subsection{Self-detection vs. cross-detection}
\label{subsec:self-detection-cross-detection}

To investigate whether LLMs can detect their own generated texts better than texts generated by other models, we compared self-detection with cross-detection. For a fair comparison, we compared each detector's performance on its own outputs with its performance on outputs generated by other models from the same generation. This controls for differences in detection difficulty across generator generations. Thereby, we averaged self- and cross-detection scores presented in Table~\ref{tab:full-results} by generations and report them in Table~\ref{tab:selfcross}. 

\begin{wraptable}{r}{0.55\columnwidth}
\centering
\caption{Model-level differences between self- and matched cross-detection (percentage points).}
\label{tab:selfcross-model-deltas}
\scriptsize
\setlength{\tabcolsep}{4pt}
\begin{tabular}{cccc}
\toprule
Generation & Detector & $\Delta$F1 (pp) & $\Delta$Accuracy (pp)\\
\midrule
\multirow{2}{*}{G1}
 & GPT-3.5 Turbo Instruct & $+0.76$ & $+0.18$ \\
 & Mixtral 8x22B & $-4.00$ & $-1.85$ \\
\midrule
\multirow{5}{*}{G2}
 & \textbf{DeepSeek-V3} & $\mathbf{-11.15}$ & $\mathbf{-8.15}$ \\
 & GLM-4-32B & $-3.17$ & $-2.06$ \\
 & GPT-4o & $-0.98$ & $-0.99$ \\
 & \textbf{Llama 3.3 70B} & $\mathbf{+2.99}$ & $\mathbf{+2.96}$ \\
 & Qwen2-72B & $+0.06$ & $+0.01$ \\
\midrule
\multirow{8}{*}{G3}
 & \textbf{Claude Opus 4.7} & $\mathbf{+0.89}$ & $\mathbf{+0.86}$ \\
 & DeepSeek-V4-Pro & $+0.30$ & $+0.05$ \\
 & \textbf{Gemini 3.1 Pro Preview} & $\mathbf{-5.40}$ & $\mathbf{-5.06}$ \\
 & \textbf{GLM-5.1} & $\mathbf{-8.43}$ & $\mathbf{-5.76}$ \\
 & GPT-5.5 & $+0.10$ & $+0.10$ \\
 & Kimi K2.6 & $+0.36$ & $+0.35$ \\
 & \textbf{MiMo V2.5 Pro} & $\mathbf{+8.00}$ & $\mathbf{+5.90}$ \\
 & \textbf{Qwen3.6 Max Preview} & $\mathbf{+1.93}$ & $\mathbf{+1.86}$ \\
\bottomrule
\end{tabular}
\end{wraptable}
For readability, we use G$x$--G$y$ cross-detection to represent detectors from G$x$ evaluating texts generated by G$y$ models. For example, G1--G2 represents G1 detectors evaluating texts generated by G2 models. For G1 detectors, self-detection achieves an F1-score of 20.8\% and an accuracy value of 51.1\%, compared with 22.4\% and 51.9\% for matched G1--G1 cross-detection. A similar pattern appears for G2 detectors comparing with the matched G2--G2 cross-detection, where self-detection reaches 52.3\% in F1 and 60.6\% in accuracy, below the corresponding cross-detection results of 54.8\% and 62.3\%. For G3 detectors, the difference becomes much smaller. Taken together, the generation-level averages do not suggest that LLMs generally detect their own outputs more accurately than outputs generated by other models from the same generation. As shown in Table~\ref{tab:full-results}, we can observe further differences between individual models. For column-wise comparison, none of the 15 detectors achieves its highest F1-score or accuracy on its own generated texts. Nevertheless, nine detectors achieve higher self-detection F1-score and accuracy than their respective averages on texts generated by other models from the same generation (i.e., GPT-3.5 Turbo Instruct from G1; Llama 3.3 70B and Qwen2-72B from G2; Claude Opus 4.7, DeepSeek-V4-Pro, GPT-5.5, Kimi K2.6, MiMo V2.5 Pro, and Qwen3.6 Max Preview from G3). This is further summarised in Table~\ref{tab:selfcross-model-deltas}, where $\Delta$ represents the self-detection score minus the detector's average score on texts generated by other models from the same generation. Hence, positive values indicate higher self-detection performance, whereas negative values suggest higher matched cross-detection performance. 

To test the statistical significance of differences observed in Table~\ref{tab:selfcross-model-deltas}, we applied prompt-clustered bootstrap tests and identify statistically significant differences for seven detectors after Holm correction (i.e., see detectors in bold in Table~\ref{tab:selfcross-model-deltas}). Four detectors (i.e., Llama 3.3 70B, Claude Opus 4.7, MiMo V2.5 Pro, and Qwen3.6 Max Preview) perform significantly better on texts generated by themselves compared with the texts generated by other models within the same generation. In contrast, DeepSeek-V3, Gemini 3.1 Pro Preview, and GLM-5.1 exhibit significant worse self-detection performance than cross-detection within the same generation. To provide a more comprehensive statistical assessment, we also aggregated the detector-level results within each generation, and conducted two sensitivity analyses that compare self-detection with cross-detection across all generations (more details can be found in Appendix~\ref{subsubsec:appendix-selfcross-statistics}). Taking all results together, we did not find evidence that support a consistent self-detection advantage or disadvantage shared across LLMs. These results challenge what \cite{Ji-J2025} reported for smaller-scale experiments, suggesting that their results are likely accidental or contain errors.

\subsection{Error pattern analysis}
\label{subsec:error-pattern-analysis}

\subsubsection{False-positive and false-negative patterns}
\label{subsec:fp-fn-patterns}

To further interpret results in Table~\ref{tab:full-results}, we examined FPRs and FNRs. A false negative occurs when an LGT is classified as human-written, whereas a false positive refers when an HGT is classified as generated by an LLM. Table~\ref{tab:error-pattern-generation} presents the mean error rates across model generations. 
\begin{wraptable}{r}{0.55\columnwidth}
\centering
\caption{Error rates across model generations (\%).}
\label{tab:error-pattern-generation}
\scriptsize
\setlength{\tabcolsep}{4pt}
\begin{tabular}{ccccc}
\toprule
& \multicolumn{3}{c}{FNR by Generator Generation} & FPR on HGTs\\
\cmidrule(lr){2-4}\cmidrule(lr){5-5}
Detector Generation & G1 & G2 & G3 & Shared Corpus\\
\midrule
G1 & 84.1 & 83.3 & 81.9 & 13.0\\
G2 & 39.9 & 39.6 & 48.1 & 36.5\\
G3 & 7.2 & 6.1 & 13.4 & 3.6\\
\bottomrule
\end{tabular}
\end{wraptable}
The results show three general patterns: 1) G1 detectors have consistently high FNRs, but a much lower FPR of 13.0\%. This suggests that G1 detectors frequently classify LGTs as human-written, exhibiting a conservative ``everything is human'' tendency. 2) G2 detectors reduce the FNR to 39.6--48.1\%, but their FPR increases to 36.5\%, suggesting that they make both type of errors relatively frequently. 3) Most G3 detectors have stronger overall performance with more balanced error profile across different generator generations, although their FNR rises to 13.4\% on G3-generated texts. This observation further confirms that G3-generated texts are comparatively more difficult to detect. More detector-level examples (e.g., GPT-3.5 Turbo Instruct's human-label tendency and GPT-4o's comparatively higher FPR) showing individual error profiles are reported in Figure~\ref{fig:detector-error-profiles} and Appendix~\ref{subsubsec:appendix-fp-fn-patterns}. Overall, detector generation is more associated with the type and magnitude of classification errors, while generator generation mainly influences the rate of misclassification of treating LGTs as human-written.
 
\subsubsection{Semantic error pattern analysis}
\label{subsubsec:semantic-error}

We further analysed the natural-language explanations using an LLM-assisted analysis and a literature-guided cue-based analysis. The LLM-assisted analysis was conducted from two complementary directions. \emph{Direction 1} keeps the exact text constant and compares explanations across different detectors. Its aim is to discover how and why different detectors make wrong judgements given identical texts. \emph{Direction 2} holds the detector and the text generation prompt (i.e., topic and rules) constant and compares explanations for texts produced by different generators. It investigates whether the same detector changes its judgement or reasoning when evaluating texts generated by different LLMs on the same topic under the same rules. Then the literature-guided cue-based analysis was conducted as a complementary and reproducible check of whether recurring patterns identified by the LLM-assisted analysis can also be observed.  

We sampled 20 text generation prompts and selected two generators and two detectors from each generation, producing 140 texts and 840 detector judgements. More details such as the sampling procedure and prompts are reported in Appendix~\ref{subsubsec:appendix-semantic-error}.
  
\paragraph{LLM-assisted semantic error pattern analysis}

We used a modern LLM (GPT-5.6 Sol with medium reasoning capability) to perform the task. The LLM received the prepared dataset, including texts, predicted labels, and explanations side by side. It was instructed to distinguish between describing a textual feature and making a valid inference about the classification task, and to support each qualitative finding with identifiable records. The complete prompts for analysing Direction 1 and Direction 2 are reported in Listings~\ref{lst:direction-a-review-prompt} and \ref{lst:direction-b-review-prompt} in Appendix~\ref{subsubsec:appendix-llm}, respectively.

In Direction 1, a typical example occurs for a GPT-3.5 generated text about medical malpractice, where DeepSeek-V3 treated the details and coherence appearing in the text as evidence of HGT, whereas DeepSeek-V4-Pro interpreted such characteristics as predictable transitions and uniform tone for evidence as LGT. In Direction 2, GPT-4o labeled empathetic and conversational responses to text generated using an academic-pressure prompt as HGT, but labeled more regularly structured responses to the text generated using the same prompt as MGT. These examples show that detectors can interpret similar properties differently and can change the reasoning across texts on the same topic. Further examples are reported in Tables~\ref{tab:direction-one-semantic-examples}~\ref{tab:direction-two-semantic-examples} in Appendix~\ref{subsubsec:appendix-llm} and more detailed analyses are also reported in Appendix~\ref{subsubsec:appendix-llm}.

\paragraph{Cue-based semantic error pattern analysis}

To complement the LLM-assisted analysis, we conducted a literature-guided cue-based analysis to examine whether the recurring patterns reported in the LLM-assisted analysis can also be discovered via the exploratory and reproducible cue-based analysis, following prior studies that perform similar explanation centric analyses~\citep{Ji-J2025, Russell-J2025}. To define the cue categories, we conducted a focused review of existing studies that investigate textual characteristics and reasoning cues for distinguishing HGTs from LGTs. Based on the small-scale review, we identified eight recurring cue categories, including \emph{structure and formatting} (SF), \emph{fluency and polish} (FP), \emph{generic or abstract language} (GAL), \emph{hedging and uncertainty} (HC), \emph{personal voice} (PV), \emph{coherence and transitions} (CT), \emph{specificity and evidence} (SE), and \emph{factuality and hallucination} (FH). The complete taxonomy and lexical patterns are reported in Tables~\ref{tab:semantic-cue-literature} and~\ref{tab:semantic-cue-keywords} in Appendix~\ref{subsubsec:appendix-cue}, the detailed experiment procedure is also reported in Appendix~\ref{subsubsec:appendix-cue}. It is worth noting that these categories are derived from a focused rather than systematic literature review and should therefore be treated as literature-guided analytical categories, rather than definitive taxonomy of classification cues.

We have observed four cue categories that are particularly consistent with the recurring findings from the LLM-assisted analysis. Since the two analyses converge on these categories, we computed how frequently each cue is mentioned in true-positive and false-negative explanations. The results are reported in Table~\ref{tab:cue-analysis-results} in Appendix~\ref{subsubsec:appendix-cue}, where each value represents the percentage of explanations containing at least one predefined term associated with the corresponding category. Therefore, the percentages measure how frequently detectors refer to each cue in their explanations, rather than how frequently the corresponding textual property occurs in the evaluated texts. SF captures references to regular or formulaic organisation, while GAL captures references to broad and insufficiently distinctive coverage. Detectors commonly used these two to support LGT judgements. For instance, SF appear in 91.4\% of GPT-4o's true-positive explanations but only 33.3\% of its false-negative ones. In contrast, SE captures the use of detailed examples, technical knowledge, or apparent expertise to support HGT judgements. This cue is more frequent in false-negative than true-positive explanations for Mixtral, GPT-4O, DeepSeek-V3, and DeepSeek-V4-Pro. For instance, DeepSeek-V3 refers to ``specific examples'' and a ``detailed and structured approach'' as reasoning evidence to incorrectly classify a LGT generated by GPT-3.5 as human-written. Finally, CT corresponds to the cue-inversion pattern observed in the LLM-assisted analysis, where the same clear and logically organised structure can lead different detectors to assign opposite labels. This appears in 83.3\% of Mixtral's true-positive explanations and 85.6\% of its false-negative explanations. Overall, the cue-based results provide convergent evidence for several patterns identified by the LLM-assisted analysis, while also showing that the same cue does not consistently lead to the same classification decision.

\section{Further Discussions, Limitations and Future Work}
\label{sec:discussions-limitation-future-work}

\textbf{Revisiting the self-detection advantage}
\citet{Ji-J2025} reported a clear self-detection advantage across the six LLMs evaluated in their study. In comparison, our results covering more (15) models do not show such a consistent advantage. However, our further analysis on comparing them within the same generation found evidence that self-detection advantages exist for some LLM models. Nevertheless, self-detection advantage appears to be model- and comparison-dependent rather than a general pattern shared by all LLMs. Therefore, our study enriches and corrects results in \citep{Ji-J2025} due to our larger coverage models, clear separation of generator and detector generations, introducing matched comparisons, and adding statistical tests. \emph{Future work should systematically examine how different factors, such as model selection, prompts, affect self-detection to identify their individual and interaction effects}.

\textbf{Watermarking and self-detection}
Several LLM vendors have announced their own watermarking mechanism to make their generated content more reliably recognisable by utilising specific detection procedures and secret keys~\cite{Dathathri-S2024,Google-G2024, Anthropic-A2026b}. We conducted an exploratory review to examine whether such mechanisms are documented for the selected models in this study. As summarised in Table~\ref{tab:watermark-llms} in Appendix~\ref{subsec:appendix-discussion}, most of the selected LLMs in this study do not have watermarking enabled at the time of our experiments. Even when a provider has announced watermarking, it is not always confirmed that the specific model/endpoint used in this study applied the watermark. Moreover, models showing a self-detection advantage do not consistently associate with those models which support watermarking. This might suggest that watermarking alone cannot explain the mixed self-detection results. \emph{Future work should be conducted to test whether watermarking improves self-detection by comparing confirmed watermarked and un-watermarked outputs from the same model and endpoint using the corresponding official detector or key}.

\textbf{Implications of cross-generational evaluation}
Our cross-generational analysis offers a new perspective to evaluate binary classification performance by considering the generation of both generator and detector. Our findings suggest that detector generation is more strongly associated with performance than generator generation, although texts generated by G3 LLMs appears to be more difficult for all detectors. However, it is important to acknowledge that the G1--G3 generation definition we used is rather simplistic, so does not necessarily reflect the most appropriate classification of LLMs. \emph{Future work should conduct a systematic literature review to develop and validate a more comprehensive taxonomy of LLM generations, supported by broader evidence on model capabilities, post-training methods, interaction paradigms, and technological development}.

\textbf{Interpreting error patterns and explanations}
We analysed the error patterns from three perspectives. The analysis of FPRs and FNRs revealed distinct generational tendencies, whereas the LLM-assisted and cue-based analyses showed consistent findings on several semantic patterns. Although the cue-based analysis can verify whether predefined cues are mentioned in an explanation, it cannot determine whether the corresponding textual properties or the explanations provide valid evidence of classification. The LLM-assisted analysis compares explanations with the evaluated texts more directly, but its interpretation may be influenced by the model (i.e., GPT 5.6 Sol)'s own understanding of HGTs and LGTs. Moreover, detector explanations may be post-hoc justifications rather than accurate and reliable descriptions of how a decision was made~\citep{Ji-J2025}. \emph{Future work should therefore involve recruitment of human annotators to confirm the the findings presented in this study, report inter-annotator agreement, and compare human coding with the two automated analyses}.

\section{Conclusion}
\label{sec:conclusion}

In this paper, we present a study of evaluating 15 general-purpose LLMs from three self-defined generations as both text generators and detectors. We instructed LLMs to perform binary classification task on a corpus of 1,000 HGTs and 15,000 LGTs. We compared detection performance and analyse LLM-generated explanations from different perspectives. Our findings suggest that, despite that texts generated by G3 generators are comparatively difficult to detect, detection performance is more strongly associated with detector generation than generator generation. The comparison between self- and cross-detection has mixed and model-specific results, providing no conclusive evidence of a consistent self-detection advantage or disadvantage. The investigation of error pattern combining FPR and FNR analysis, LLM-assisted analysis, and cue-based analysis, identify different generational tendencies and show that detectors can interpret the same textual cues differently. Overall, general-purpose LLMs show great potential as detectors, but their performance and explanations remain model-dependent, and the LLM-generated explanations should be used with caution as they may contain errors.



\subsection*{Ethics statement}

This study does not involve human participants or any private personal data. All experiments were conducted using publicly available LLMs and datasets, and by ourselves as researchers.

\subsection*{Reproducibility statement}

To support reproducibility, we release a repository containing the complete HGT and LGT corpus, raw detector judgements with explanations, text generation and detection prompts, source code to perform the text generation task and binary classification task, source code to produce all results and conduct all analyses presented in the manuscript. The repository can be accessed from \url{https://github.com/hyyuan/detect-llm-generated-texts}. Section~\ref{sec:experimental-design} presents the detailed experiment design while Appendix~\ref{sec:appendix} provides all necessary materials to support further explanations of experiment design.



\bibliography{main}

@article{Bhattacharjee-A2024,
  title     = {Fighting Fire with Fire: Can {ChatGPT} Detect {AI}-Generated Text?},
  author    = {Bhattacharjee, Amrita and Liu, Huan},
  journal   = {ACM SIGKDD Explorations Newsletter},
  volume    = {25},
  number    = {2},
  pages     = {14--21},
  year      = {2024},
  publisher = {Association for Computing Machinery},
  doi       = {10.1145/3655103.3655106},
}

@inproceedings{Caiado-A2025,
  title     = {{AI} Content Self-Detection for Transformer-Based Large Language Models},
  author    = {Caiado, Ant{\^o}nio Junior Alves and Hahsler, Michael},
  booktitle = {Intelligent Systems and Applications},
  series    = {Lecture Notes in Networks and Systems},
  volume    = {1554},
  pages     = {147--168},
  year      = {2025},
  publisher = {Springer Nature Switzerland},
  doi       = {10.1007/978-3-031-99965-9_10},
}

@inproceedings{Panickssery-A2024,
  title     = {{LLM} Evaluators Recognize and Favor Their Own Generations},
  author    = {Panickssery, Arjun and Bowman, Samuel R. and Feng, Shi},
  booktitle = {Advances in Neural Information Processing Systems},
  volume    = {37},
  year      = {2024},
  doi       = {10.52202/079017-2197},
}

@article{Bai-X2025,
  title   = {Know Thyself? On the Incapability and Implications of {AI} Self-Recognition},
  author  = {Bai, Xiaoyan and Shrivastava, Aryan and Holtzman, Ari and Tan, Chenhao},
  journal = {arXiv preprint arXiv:2510.03399},
  year    = {2025},
  doi     = {10.48550/arXiv.2510.03399},
}

@article{StAmand-J2026,
  title   = {Self-Generated Text Recognition: Quality Heuristics, Cross-Task Transfer, and Downstream Bias in {LLM} Evaluation},
  author  = {{St. Amand}, Jesse and Canavan, Callum and Imran, Sohaib and Hewson, Joseph and Lutz, Aaron and Feng, Shi and Radmard, Puria and Wells, Lennie},
  journal = {arXiv preprint arXiv:2608.26159},
  year    = {2026},
  doi     = {10.48550/arXiv.2608.26159},
}

@inproceedings{Czybik-S2026,
  title     = {A Large-Scale Study of Personalized Phishing Using Large Language Models},
  author    = {Czybik, Stefan and Kouam, Anne Josiane and Heubl, Peter and Nold, Jan Magnus and Rieck, Konrad},
  booktitle = {35th USENIX Security Symposium (USENIX Security 26)},
  year      = {2026},
  publisher = {USENIX Association},
  month     = aug,
  url       = {https://www.usenix.org/conference/usenixsecurity26/presentation/czybik},
}

@inproceedings{Dugan-L2024,
  title     = {{RAID}: A Shared Benchmark for Robust Evaluation of Machine-Generated Text Detectors},
  author    = {Dugan, Liam and Hwang, Alyssa and Trhl{\'i}k, Filip and Zhu, Andrew and Ludan, Josh Magnus and Xu, Hainiu and Ippolito, Daphne and Callison-Burch, Chris},
  booktitle = {Proceedings of the 62nd Annual Meeting of the Association for Computational Linguistics (Volume 1: Long Papers)},
  pages     = {12463--12492},
  year      = {2024},
  publisher = {Association for Computational Linguistics},
  doi       = {10.18653/v1/2024.acl-long.674},
}

@inproceedings{Guo-R2026,
  title     = {{HLD}: Approximate Hierarchical Linguistic Distribution Modeling for {LLM}-Generated Text Detection},
  author    = {Guo, Rui and Zeng, Weibin and Wu, Fuzhang and Kong, Yan and Shen, Sicheng and Wu, Yanjun and Dong, Weiming},
  booktitle = {The Fourteenth International Conference on Learning Representations},
  year      = {2026},
  url       = {https://openreview.net/forum?id=l9mqzHROGu},
}

@inproceedings{Hans-A2024,
  title     = {Spotting {LLM}s with Binoculars: Zero-Shot Detection of Machine-Generated Text},
  author    = {Hans, Abhimanyu and Schwarzschild, Avi and Cherepanova, Valeriia and Kazemi, Hamid and Saha, Aniruddha and Goldblum, Micah and Geiping, Jonas and Goldstein, Tom},
  booktitle = {Proceedings of the 41st International Conference on Machine Learning},
  volume    = {235},
  series    = {Proceedings of Machine Learning Research},
  pages     = {17519--17537},
  year      = {2024},
  publisher = {PMLR},
  url       = {https://proceedings.mlr.press/v235/hans24a.html},
}

@inproceedings{Huang-Y2025,
  title     = {{MAGRET}: Machine-Generated Text Detection with Rewritten Texts},
  author    = {Huang, Yifei and Cao, Jiuxin and Luo, Hanyu and Guan, Xin and Liu, Bo},
  booktitle = {Proceedings of the 31st International Conference on Computational Linguistics},
  pages     = {8336--8346},
  year      = {2025},
  publisher = {Association for Computational Linguistics},
  url       = {https://aclanthology.org/2025.coling-main.557/},
}

@article{Ji-J2024,
  title   = {Detecting Machine-Generated Texts: Not Just ``{AI} vs Humans'' and Explainability Is Complicated},
  author  = {Ji, Jiazhou and Li, Ruizhe and Li, Shujun and Guo, Jie and Qiu, Weidong and Huang, Zheng and Chen, Chiyu and Jiang, Xiaoyu and Lu, Xinru},
  journal = {arXiv preprint arXiv:2406.18259},
  year    = {2024},
  doi     = {10.48550/arXiv.2406.18259},
}

@article{Ji-J2025,
  title   = {{``I Know Myself Better, but Not Really Greatly''}: How Well Can {LLM}s Detect and Explain {LLM}-Generated Texts?},
  author  = {Ji, Jiazhou and Guo, Jie and Qiu, Weidong and Huang, Zheng and Xu, Yang and Lu, Xinru and Jiang, Xiaoyu and Li, Ruizhe and Li, Shujun},
  journal = {arXiv preprint arXiv:2502.12743},
  year    = {2025},
  doi     = {10.48550/arXiv.2502.12743},
}

@inproceedings{Russell-J2025,
  title     = {People Who Frequently Use {ChatGPT} for Writing Tasks Are Accurate and Robust Detectors of {AI}-Generated Text},
  author    = {Russell, Jenna and Karpinska, Marzena and Iyyer, Mohit},
  booktitle = {Proceedings of the 63rd Annual Meeting of the Association for Computational Linguistics (Volume 1: Long Papers)},
  pages     = {5342--5373},
  year      = {2025},
  publisher = {Association for Computational Linguistics},
  doi       = {10.18653/v1/2025.acl-long.267},
}

@inproceedings{Kirchenbauer-J2023,
  title     = {A Watermark for Large Language Models},
  author    = {Kirchenbauer, John and Geiping, Jonas and Wen, Yuxin and Katz, Jonathan and Miers, Ian and Goldstein, Tom},
  booktitle = {Proceedings of the 40th International Conference on Machine Learning},
  volume    = {202},
  series    = {Proceedings of Machine Learning Research},
  pages     = {17061--17084},
  year      = {2023},
  publisher = {PMLR},
  url       = {https://proceedings.mlr.press/v202/kirchenbauer23a.html},
}

@article{Kovacs-B2024,
  title   = {The Turing Test of Online Reviews: Can We Tell the Difference Between Human-Written and {GPT-4}-Written Online Reviews?},
  author  = {Kov{\'a}cs, Bal{\'a}zs},
  journal = {Marketing Letters},
  volume  = {35},
  number  = {4},
  pages   = {651--666},
  year    = {2024},
  doi     = {10.1007/s11002-024-09729-3},
}

@inproceedings{Layton-S2026,
  title     = {{AI} Wrote My Paper and All I Got Was This False Negative: Measuring the Efficacy of Commercial {AI} Text Detectors},
  author    = {Layton, Seth and Medeiros, Bernardo B. P. and Butler, Kevin and Traynor, Patrick},
  booktitle = {2026 IEEE Symposium on Security and Privacy},
  pages     = {213--232},
  year      = {2026},
  publisher = {IEEE},
  doi       = {10.1109/SP63933.2026.00088},
}

@inproceedings{Lee-J2023,
  title     = {Do Language Models Plagiarize?},
  author    = {Lee, Jooyoung and Le, Thai and Chen, Jinghui and Lee, Dongwon},
  booktitle = {Proceedings of the ACM Web Conference 2023},
  series    = {WWW '23},
  pages     = {3637--3647},
  year      = {2023},
  publisher = {Association for Computing Machinery},
  doi       = {10.1145/3543507.3583199},
}

@article{Leinonen-J2026,
  title   = {Distinguishing Artificial from Authentic: Evaluating {LLM}s for Detecting {LLM}-Generated Content},
  author  = {Leinonen, Juho and Denny, Paul},
  journal = {arXiv preprint arXiv:2607.20446},
  year    = {2026},
  doi     = {10.48550/arXiv.2607.20446},
}

@article{Liang-W2023,
  title   = {{GPT} Detectors Are Biased Against Non-Native English Writers},
  author  = {Liang, Weixin and Yuksekgonul, Mert and Mao, Yining and Wu, Eric and Zou, James},
  journal = {Patterns},
  volume  = {4},
  number  = {7},
  pages   = {100779},
  year    = {2023},
  doi     = {10.1016/j.patter.2023.100779},
}

@inproceedings{Li-J2024,
  title     = {Fundamental Capabilities of Large Language Models and Their Applications in Domain Scenarios: A Survey},
  author    = {Li, Jiawei and Yang, Yizhe and Bai, Yu and Zhou, Xiaofeng and Li, Yinghao and Sun, Huashan and Liu, Yuhang and Si, Xingpeng and Ye, Yuhao and Wu, Yixiao and Lin, Yiguan and Xu, Bin and Ren, Bowen and Feng, Chong and Gao, Yang and Huang, Heyan},
  booktitle = {Proceedings of the 62nd Annual Meeting of the Association for Computational Linguistics (Volume 1: Long Papers)},
  pages     = {11116--11141},
  year      = {2024},
  publisher = {Association for Computational Linguistics},
  doi       = {10.18653/v1/2024.acl-long.599},
}

@inproceedings{Mitchell-E2023,
  title     = {{DetectGPT}: Zero-Shot Machine-Generated Text Detection Using Probability Curvature},
  author    = {Mitchell, Eric and Lee, Yoonho and Khazatsky, Alexander and Manning, Christopher D. and Finn, Chelsea},
  booktitle = {Proceedings of the 40th International Conference on Machine Learning},
  volume    = {202},
  series    = {Proceedings of Machine Learning Research},
  pages     = {24950--24962},
  year      = {2023},
  publisher = {PMLR},
  url       = {https://proceedings.mlr.press/v202/mitchell23a.html},
}

@inproceedings{Pedrotti-A2025,
  title     = {Stress-Testing Machine Generated Text Detection: Shifting Language Models Writing Style to Fool Detectors},
  author    = {Pedrotti, Andrea and Papucci, Michele and Ciaccio, Cristiano and Miaschi, Alessio and Puccetti, Giovanni and Dell'Orletta, Felice and Esuli, Andrea},
  booktitle = {Findings of the Association for Computational Linguistics: ACL 2025},
  pages     = {3010--3031},
  year      = {2025},
  publisher = {Association for Computational Linguistics},
  doi       = {10.18653/v1/2025.findings-acl.156},
}

@inproceedings{Su-Z2025,
  title     = {{HACo-Det}: A Study Towards Fine-Grained Machine-Generated Text Detection under Human--{AI} Coauthoring},
  author    = {Su, Zhixiong and Wang, Yichen and Wan, Herun and Zhang, Zhaohan and Luo, Minnan},
  booktitle = {Proceedings of the 63rd Annual Meeting of the Association for Computational Linguistics (Volume 1: Long Papers)},
  pages     = {22015--22036},
  year      = {2025},
  publisher = {Association for Computational Linguistics},
  doi       = {10.18653/v1/2025.acl-long.1069},
}

@inproceedings{Sun-Y2026,
  title     = {{D\&R}: Recovery-Based {AI}-Generated Text Detection via a Single Black-Box {LLM} Call},
  author    = {Sun, Yuxia and Zhang, Ran and Sun, Aoxiang and Li, Xu and Liu, Zitao and Guo, Jingcai},
  booktitle = {The Fourteenth International Conference on Learning Representations},
  year      = {2026},
  url       = {https://openreview.net/forum?id=FiMZSxo4DO},
}

@inproceedings{Tufts-B2025,
  title     = {A Practical Examination of {AI}-Generated Text Detectors for Large Language Models},
  author    = {Tufts, Brian and Zhao, Xuandong and Li, Lei},
  booktitle = {Findings of the Association for Computational Linguistics: NAACL 2025},
  pages     = {4839--4856},
  year      = {2025},
  publisher = {Association for Computational Linguistics},
  doi       = {10.18653/v1/2025.findings-naacl.271},
}

@inproceedings{Wang-Y2024,
  title     = {{M4GT-Bench}: Evaluation Benchmark for Black-Box Machine-Generated Text Detection},
  author    = {Wang, Yuxia and Mansurov, Jonibek and Ivanov, Petar and Su, Jinyan and Shelmanov, Artem and Tsvigun, Akim and {Mohammed Afzal}, Osama and Mahmoud, Tarek and Puccetti, Giovanni and Arnold, Thomas and Aji, Alham Fikri and Habash, Nizar and Gurevych, Iryna and Nakov, Preslav},
  booktitle = {Proceedings of the 62nd Annual Meeting of the Association for Computational Linguistics (Volume 1: Long Papers)},
  pages     = {3964--3992},
  year      = {2024},
  publisher = {Association for Computational Linguistics},
  doi       = {10.18653/v1/2024.acl-long.218},
}

@inproceedings{Vykopal-I2024,
  title     = {Disinformation Capabilities of Large Language Models},
  author    = {Vykopal, Ivan and Pikuliak, Mat{\'u}{\v{s}} and Srba, Ivan and Moro, Robert and Macko, Dominik and Bielikov{\'a}, Maria},
  booktitle = {Proceedings of the 62nd Annual Meeting of the Association for Computational Linguistics (Volume 1: Long Papers)},
  pages     = {14830--14847},
  year      = {2024},
  publisher = {Association for Computational Linguistics},
  doi       = {10.18653/v1/2024.acl-long.793},
}

@article{Wu-J2025a,
  title     = {A Survey on {LLM}-Generated Text Detection: Necessity, Methods, and Future Directions},
  author    = {Wu, Junchao and Yang, Shu and Zhan, Runzhe and Yuan, Yulin and Chao, Lidia Sam and Wong, Derek Fai},
  journal   = {Computational Linguistics},
  volume    = {51},
  number    = {1},
  pages     = {275--338},
  year      = {2025},
  doi       = {10.1162/coli_a_00549},
}

@inproceedings{Wu-C2026,
  title     = {Beyond Raw Detection Scores: Markov-Informed Calibration for Boosting Machine-Generated Text Detection},
  author    = {Wu, Chenwang and Cheung, Yiu-ming and Zhang, Shuhai and Han, Bo and Lian, Defu},
  booktitle = {The Fourteenth International Conference on Learning Representations},
  year      = {2026},
  url       = {https://openreview.net/forum?id=Lzwkeg2o2z},
}

@inproceedings{Wu-J2025b,
  title     = {{MoSEs}: Uncertainty-Aware {AI}-Generated Text Detection via Mixture of Stylistics Experts with Conditional Thresholds},
  author    = {Wu, Junxi and Wang, Jinpeng and Liu, Zheng and Chen, Bin and Hu, Dongjian and Wu, Hao and Xia, Shu-Tao},
  booktitle = {Proceedings of the 2025 Conference on Empirical Methods in Natural Language Processing},
  pages     = {5786--5805},
  year      = {2025},
  publisher = {Association for Computational Linguistics},
  doi       = {10.18653/v1/2025.emnlp-main.294},
}

@inproceedings{Xiong-R2026,
  title     = {Verifiable {LLM}-Generated Text Detection via Projected Semantic-Structural Distributions},
  author    = {Xiong, Ruochong and Li, Qien and Lian, Wangwang and Wan, Yulong and Xue, Hanlin and Tan, Zhouxing and Yang, Han and Lu, Fengyu and Liu, Junfei},
  booktitle = {Proceedings of the 64th Annual Meeting of the Association for Computational Linguistics (Volume 1: Long Papers)},
  pages     = {14005--14042},
  year      = {2026},
  publisher = {Association for Computational Linguistics},
  doi       = {10.18653/v1/2026.acl-long.638},
}

@inproceedings{Luo-H2026,
  title     = {{SpecDetect}: Simple, Fast, and Training-Free Detection of {LLM}-Generated Text via Spectral Analysis},
  author    = {Luo, Haitong and Zhang, Weiyao and Wang, Suhang and Zou, Wenji and Lin, Chungang and Meng, Xuying and Zhang, Yujun},
  booktitle = {Proceedings of the AAAI Conference on Artificial Intelligence},
  volume    = {40},
  number    = {38},
  pages     = {32356--32364},
  year      = {2026},
  doi       = {10.1609/aaai.v40i38.40510},
}

@inproceedings{Zhou-H2026,
  title     = {Learn-to-Distance: Distance Learning for Detecting {LLM}-Generated Text},
  author    = {Zhou, Hongyi and Zhu, Jin and Ye, Kai and Yang, Ying and Xu, Erhan and Shi, Chengchun},
  booktitle = {The Fourteenth International Conference on Learning Representations},
  year      = {2026},
  url       = {https://openreview.net/forum?id=2ZUPeEM3FH},
}

@inproceedings{Clark-E2021,
  title     = {All That's `Human' Is Not Gold: Evaluating Human Evaluation of Generated Text},
  author    = {Clark, Elizabeth and August, Tal and Serrano, Sofia and Haduong, Nikita and Gururangan, Suchin and Smith, Noah A.},
  booktitle = {Proceedings of the 59th Annual Meeting of the Association for Computational Linguistics and the 11th International Joint Conference on Natural Language Processing (Volume 1: Long Papers)},
  pages     = {7282--7296},
  year      = {2021},
  publisher = {Association for Computational Linguistics},
  doi       = {10.18653/v1/2021.acl-long.565},
}

@inproceedings{Ippolito-D2020,
  title     = {Automatic Detection of Generated Text Is Easiest When Humans Are Fooled},
  author    = {Ippolito, Daphne and Duckworth, Daniel and Callison-Burch, Chris and Eck, Douglas},
  booktitle = {Proceedings of the 58th Annual Meeting of the Association for Computational Linguistics},
  pages     = {1808--1822},
  year      = {2020},
  publisher = {Association for Computational Linguistics},
  doi       = {10.18653/v1/2020.acl-main.164},
}

@article{Mitrovic-S2023,
  title   = {{ChatGPT} or Human? Detect and Explain: Explaining Decisions of Machine Learning Models for Detecting Short {ChatGPT}-Generated Text},
  author  = {Mitrovi{\'c}, Sandra and Andreoletti, Davide and Ayoub, Omran},
  journal = {arXiv preprint arXiv:2301.13852},
  year    = {2023},
  doi     = {10.48550/arXiv.2301.13852},
}

@article{Ji-Z2023,
  title     = {Survey of Hallucination in Natural Language Generation},
  author    = {Ji, Ziwei and Lee, Nayeon and Frieske, Rita and Yu, Tiezheng and Su, Dan and Xu, Yan and Ishii, Etsuko and Bang, Yejin and Madotto, Andrea and Fung, Pascale},
  journal   = {ACM Computing Surveys},
  volume    = {55},
  number    = {12},
  pages     = {248:1--248:38},
  year      = {2023},
  publisher = {Association for Computing Machinery},
  doi       = {10.1145/3571730},
}

@article{Dathathri-S2024,
  title   = {Scalable Watermarking for Identifying Large Language Model Outputs},
  author  = {Dathathri, Sumanth and See, Abigail and Ghaisas, Sumedh and Huang, Po-Sen and McAdam, Rob and Welbl, Johannes and Bachani, Vandana and Kaskasoli, Alex and Stanforth, Robert and Matejovicova, Tatiana and Hayes, Jamie and Vyas, Nidhi and Al Merey, Majd and Brown-Cohen, Jonah and Bunel, Rudy and Balle, Borja and Cemgil, Taylan and Ahmed, Zahra and Stacpoole, Kitty and Shumailov, Ilia and Baetu, Ciprian and Gowal, Sven and Hassabis, Demis and Kohli, Pushmeet},
  journal = {Nature},
  volume  = {634},
  number  = {8035},
  pages   = {818--823},
  year    = {2024},
  doi     = {10.1038/s41586-024-08025-4},
}

@misc{Google-G2024,
  title        = {Watermarking {AI}-Generated Text and Video with {SynthID}},
  author       = {{Google DeepMind}},
  year         = {2024},
  howpublished = {Google DeepMind},
  url          = {https://deepmind.google/blog/watermarking-ai-generated-text-and-video-with-synthid/},
}

@misc{Anthropic-A2026b,
  title        = {How {Claude}'s Text Watermark Works},
  author       = {{Anthropic}},
  year         = {2026},
  howpublished = {Anthropic},
  url          = {https://www.anthropic.com/news/claude-text-watermark},
}

@misc{Mistral-M2024a,
  title        = {Cheaper, Better, Faster, Stronger: {Mixtral} 8x22B},
  author       = {{Mistral AI}},
  year         = {2024},
  howpublished = {Mistral AI},
  url          = {https://mistral.ai/news/mixtral-8x22b/},
}

@misc{OpenAI-O2023,
  title        = {{GPT-3.5 Turbo Instruct} Model Documentation},
  author       = {{OpenAI}},
  year         = {2023},
  howpublished = {OpenAI API Documentation},
  url          = {https://developers.openai.com/api/docs/models/gpt-3.5-turbo-instruct},
}

@misc{ZAI-Z2025,
  title        = {{GLM-4-32B-0414} Model Card},
  author       = {{Z.AI}},
  year         = {2025},
  howpublished = {Hugging Face Model Card},
  url          = {https://huggingface.co/zai-org/GLM-4-32B-0414},
}

@misc{OpenAI-O2024,
  title        = {{GPT-4o} System Card},
  author       = {{OpenAI}},
  year         = {2024},
  howpublished = {OpenAI},
  url          = {https://openai.com/index/gpt-4o-system-card/},
}

@article{Qwen-Q2024,
  title   = {{Qwen2.5} Technical Report},
  author  = {{Qwen Team}},
  journal = {arXiv preprint arXiv:2412.15115},
  year    = {2024},
  doi     = {10.48550/arXiv.2412.15115},
}

@article{DeepSeek-D2024,
  title   = {{DeepSeek-V3} Technical Report},
  author  = {{DeepSeek-AI}},
  journal = {arXiv preprint arXiv:2412.19437},
  year    = {2024},
  doi     = {10.48550/arXiv.2412.19437},
}

@misc{Meta-M2024,
  title        = {{Llama 3.3 70B Instruct} Model Card},
  author       = {{Meta AI}},
  year         = {2024},
  howpublished = {Hugging Face Model Card},
  url          = {https://huggingface.co/meta-llama/Llama-3.3-70B-Instruct},
}

@misc{OpenAI-O2026,
  title        = {Introducing {GPT-5.5}},
  author       = {{OpenAI}},
  year         = {2026},
  howpublished = {OpenAI},
  url          = {https://openai.com/index/introducing-gpt-5-5/},
}

@misc{Anthropic-A2026,
  title        = {Introducing {Claude Opus 4.7}},
  author       = {{Anthropic}},
  year         = {2026},
  howpublished = {Anthropic},
  url          = {https://www.anthropic.com/news/claude-opus-4-7},
}

@misc{Google-G2026,
  title        = {{Gemini 3.1 Pro}: A Smarter Model for Your Most Complex Tasks},
  author       = {{Google}},
  year         = {2026},
  howpublished = {Google},
  url          = {https://blog.google/innovation-and-ai/models-and-research/gemini-models/gemini-3-1-pro/},
}

@misc{DeepSeek-D2026,
  title        = {{DeepSeek-V4} Preview Release},
  author       = {{DeepSeek-AI}},
  year         = {2026},
  howpublished = {DeepSeek API Documentation},
  url          = {https://api-docs.deepseek.com/news/news260424/},
}

@misc{Qwen-Q2026,
  title        = {{Qwen3.6-Max-Preview}},
  author       = {{Qwen Team}},
  year         = {2026},
  howpublished = {Qwen},
  url          = {https://qwen.ai/blog?id=qwen3.6-max-preview},
}

@misc{ZAI-Z2026,
  title        = {{GLM-5.1}: Overview},
  author       = {{Z.AI}},
  year         = {2026},
  howpublished = {Z.AI Developer Documentation},
  url          = {https://docs.z.ai/guides/llm/glm-5.1},
}

@misc{Xiaomi-X2026,
  title        = {{Xiaomi MiMo-V2.5} Series Open-Sourced},
  author       = {{Xiaomi MiMo Team}},
  year         = {2026},
  howpublished = {Xiaomi MiMo},
  url          = {https://mimo.mi.com/docs/en-US/news/latest/v2.5-open-sourced},
}

@misc{Moonshot-M2026,
  title        = {{Kimi K2.6}: Leading Open-Source Model in Coding and Agent Tasks},
  author       = {{Moonshot AI}},
  year         = {2026},
  howpublished = {Kimi},
  url          = {https://www.kimi.ai/ai-models/kimi-k2-6},
}

@misc{OpenRouter-O2026,
  title        = {OpenRouter Quickstart Guide},
  author       = {{OpenRouter}},
  year         = {2026},
  howpublished = {OpenRouter Documentation},
  url          = {https://openrouter.ai/docs/quickstart},
}
\bibliographystyle{iclr2027_conference}

\appendix

\section{Appendix}
\label{sec:appendix}

\subsection{Supporting materials for Section~\ref{sec:related-work}}
\label{subsec:appendix-related-work}

This section presents a more detailed review on related work.

\paragraph{LLM-generated text detection}

Prior work conceptualised LGTs detection as a binary classification task to distinguish LGTs from HGTs and has developed various methods to facilitate the detection, including statistics-based detectors, neural-based detectors, and watermarking approaches~\cite{Wu-J2025a}. Previous study found that supervised classifiers often perform well when evaluated on texts from the same domains and generators as their training data~\cite{Wang-Y2024, Wu-J2025a}. However, evaluation on M4GT-Bench and RAID show that detectors' performance could vary substantially on unfamiliar data, particularly when the generator, domain, decoding setting changes~\cite{Wang-Y2024, Dugan-L2024}. 

Instead, researchers have been exploring various training-free approaches to facilitate LGTs detection. DetectGPT exploits local probability curvature under perturbations to support LGTs detection~\cite{Mitchell-E2023}. Binoculars was proposed as a novel LLM detector that compares perplexity estimates using a pair of pre-trained LLMs~\cite{Hans-A2024}. More recently, SpecDetect takes the signal processing perspective to detect frequency-domain differences in token log-probability sequences~\cite{Luo-H2026}. Rewriting-based approaches use an LLM's response to an input as a provenance signal. MAGRET compares an input with LLM-generated rewrite~\cite{Huang-Y2025}. D\&R analyses the input and measures how well an LLM can recover its semantic and structural properties~\cite{Sun-Y2026}. Learn-to-Distance learns adaptive metrics between original texts and rewritten texts to help detect LGTs~\cite{Zhou-H2026}.

Beyond probability and rewriting based approaches reviewed in above paragraphs, researchers have explored a broader range of textual features such as writing style, token-level dependencies, syntax, and document structure, to facilitate LGTs detection. MoSEs utilises style models and makes detection decision based on an adaptive threshold that is associated with the characteristics and uncertainty of each input~\cite{Wu-J2025b}. HLD-Detector compares HGTs and LGTs at multiple linguistic levels in terms of word choice, syntactic structure, and meanings~\cite{Guo-R2026}. ProSSD is a statistical framework that learns compact semantic features conditioned on syntactic structure and can distinguish between HGTs and LGTs using a likelihood-ratio metric that combines Mahalanobis and Wasserstein distances~\cite{Xiong-R2026}. Moreover, Wu et al. identified two recurring properties of contextual detection scores: Neighbor Similarity and Initial Instability. Subsequently, Markov-informed score calibration strategy is proposed to model these relationships with a Markov random field and use a lightweight mean-field approximation to calibrate the raw scores, allowing the method to be integrated to existing detectors~\cite{Wu-C2026}. Nevertheless, detection reliability remains sensitive to evaluation settings. In a study that conducts broad evaluations, Turfs et al. showed that detector performance varies across unseen dataset, domains, and generators. They also indicated that strong aggregate metrics can obscure weak performance at low false-positive rates~\cite{Tufts-B2025}. Similar limitations are also found in a study that investigates commercial systems, Layton et al. reported significant differences in false-positive and false-negative rates across detectors. They show that simple change of styles of prompts, such as using more complex vocabulary, can reduce detection accuracy~\cite{Layton-S2026}. 

\paragraph{LLMs as detectors and self-detection}

Complementing the above approaches, another line of research looks at using general-purpose LLMs themselves as detectors, prompting them in a zero-shot setting to classify a given text as either HGT or LGT. Unlike many statistical and likelihood-based methods mentioned above, this approach does not require access to token probabilities, but it can still make a binary HGT--LGT classification decision for a given text as well as providing a natural-language explanation. Bhattacharjee and Liu evaluated ChatGPT as a zero-shot detector and found strongly asymmetric performance between HGTs and LGTs. GPT-3.5 can largely recognise human-written articles but fail to identify most machine-generated articles. In comparison, GPT-4 classifies almost all texts as machine-generated~\cite{Bhattacharjee-A2024}. The study that most closely related to ours is from Ji et al., who systematically evaluated six open- and closed-source LLMs as both generators and detectors under binary and ternary classification settings~\cite{Ji-J2025}. Across their experiments, they discovered the self-detection advantage where an LLM generally identifies texts generated by itself more accurately than texts generated by other LLMs (i.e., cross-detection).

There are other studies looking at self-detection and self-recognition of LLMs. Panickssery et al. found that GPT-4 and Llama 2 can identify text generated by themselves among texts generated by other LLMs or humans at above-chance rate. They further suggested that the stronger self-recognition is related to so-called self-preference, meaning that LLM would rate its own generation more favourably than those generated from other models~\cite{Panickssery-A2024}. Differently, Bai et al. evaluated ten LLMs and found that self-recognition is rarely above chance. Moreover, the selected models in their study also seems to attribute high-quality text to the GPT and Claude families regardless of its actual source~\cite{Bai-X2025}. In a another small scale comparison study, Caido and Hahslef had inconsistent findings, where Bard and ChatGPT exhibit strong self-detection but not Claude~\cite{Caiado-A2025}. More recently, using nested sets of 13, 18, or 21 models, depending on their compatibility with each experimental condition, St. Amand et al. showed that the measured self-recognition varies significantly under different contexts, such as evaluation format, conversation structure, and task domain~\cite{StAmand-J2026}. Consistent with this context dependence, Leinonen and Denny found that self-detection performance can be affected by prompt framing, response length, and stylistic instructions in a study that uses GPT-4o as both the text generator and detector~\cite{Leinonen-J2026}. 

\paragraph{Error analysis and explanations}

In addition to instructing LLMs to judge if a given text is a LGT or a HGT, some studies prompt LLM detectors to explain their judgements in natural language, allowing researchers to learn the underlying textual cues and reasoning~\cite{Ji-J2025,Russell-J2025}. However, based on existing studies, these explanations do not necessarily provide a complete or reliable account of why the detectors made their decisions. Ji et al. examined detector classification results and their explanations collectively and discovered that an LLM can reach the correct decision for the wrong reason. They also identified three recurring issues, including using weak features (e.g., polished grammar) as reliable evidence to distinguish HGTs from LGTs, using features that are not present in the text, and using genuine features but interpreting their relevance incorrectly~\cite{Ji-J2025}. 

In addition to studies that focus on analysing LLM-generated explanations, Russell et al. collected 1500 free-text explanations from five annotators experienced in using LLMs for writing tasks. They developed a coding scheme for recurring detection cues and applied it with GPT-4o. The analysis shows that the annotators frequently refer to vocabulary, sentence structure, grammar, and originality as cues to make judgments. By comparing explanations associated with correct and incorrect decisions, they further demonstrated that the same cue can be useful or misleading. For instance, ``AI-like'' words, such as , \emph{delve} and \emph{crucial}, can contribute to false positives, whereas contractions and colloquial language cause LGTs to be misclassified as human-written~\cite{Russell-J2025}. Moreover, researchers have also examined the explanations offered by commercially available detectors. Ji et al., conducted a study to compare the metrics reported by GPTZero with explanations written by human annotators. They reported that cues such as fluency, linguistic errors, repetition, perplexity, and readability can overlap between HGTs and LGTs. In addition, they indicated that the metrics used by GPTZero can only provide limited insights into how GPTZero reached its final decision~\cite{Ji-J2024}.

\paragraph{Research gaps} Taken together, we have identified a few research gaps that motivated us to carry out this study. Systematic evaluations of general-purpose LLMs acting as both text generators and detectors remain limited in scale. The study most closely related to ours evaluates only six models~\cite{Ji-J2025}, leaving it unclear how self- and cross-detection would change across a broader set of models and generations under the same experimental setting. Moreover, prior research has examined LLM explanations, human interpretations, and commercial detector metrics sparsely in separate settings. However, it remains unclear about how the cues reported by general-purpose LLM detectors relate to correct decisions, false positives, and false negatives across self- and cross-detection settings.

\subsection{Supporting materials for Section~\ref{sec:experimental-design}}
\label{subsec:appendix-curpus-construction}

\subsubsection{Corpus construction}
\label{subsubsec:appendix-curpus-construction}

\begin{lstlisting}[language=json,caption={An example of the matched text generation prompt used to generate LGTs},label={lst:generation-prompt-example}] 
Discuss the properties of gases according to the ideal gas law. How does temperature, pressure, and volume influence gas behavior? Answer in 150--300 words.
\end{lstlisting}

\begin{lstlisting}[language=json,caption={Example of the generated text using GPT-5.5},label={lst:generated_text_example}] 
The ideal gas law describes how gases behave when they are assumed to consist of tiny particles that move freely, have negligible volume, and do not attract or repel one another. It states that pressure times volume is equal to the amount of gas times the gas constant times absolute temperature. This relationship helps explain how a change in one property affects the others when the amount of gas is fixed.
Temperature measures the average kinetic energy of gas particles. When temperature increases, particles move faster and collide more forcefully with the walls of their container. If volume is held constant, this raises the pressure. If pressure is held constant, the gas expands and its volume increases. Temperature must be measured in kelvins because gas behavior depends on absolute temperature.
Pressure is caused by gas particles colliding with surfaces. If a gas is compressed into a smaller volume while temperature stays the same, particles collide with the container walls more often, so pressure increases. This is why pressure and volume are inversely related at constant temperature.
Volume is the space occupied by the gas. Increasing volume gives particles more room to move, reducing collision frequency and lowering pressure if temperature does not change. Although real gases can deviate from ideal behavior, especially at high pressure or low temperature, the ideal gas law is a useful model for predicting gas behavior in many everyday situations.
\end{lstlisting}

\subsubsection{Binary classification}
\label{subsubsec:appendix-binary-classification}

\begin{lstlisting}[language=json,caption={Instruction template used for binary detection.},label={lst:binary_prompt}] 
Please determine whether the following text is generated by large language models or by a human, and provide a detailed explanation for your decision. You must respond in strict JSON format. The value of "answer" must be exactly one of: "human" or "machine"---no other values, capitalisation, or punctuation are allowed. Structure your answer as follows: {"answer": "human" or "machine", "explanation": "your explanation here"}
\end{lstlisting}

\begin{lstlisting}[language=json,caption={Example of self-detection response using GPT-5.5 as both generator and detector},label={lst:binary_response_example}] 
"id": 0,
"source": "gpt-5.5",
"ground_truth": "machine",
"response": {"answer":"machine",
"explanation":"The text is highly polished, generic, and evenly structured in a way that is characteristic of large language model output. It explains the ideal gas law in broad educational terms without personal perspective, unusual phrasing, mistakes, or specific contextual details. The paragraphs follow a predictable pattern: define a concept, explain the mechanism, then state the consequence. The wording is clear but somewhat formulaic, with repeated explanatory transitions such as "If volume is held constant," "If pressure is held constant," and "This is why." While a human could certainly write this, the lack of distinctive style, examples, or variation, combined with the smooth textbook-like completeness, makes it more likely to be machine-generated."}
\end{lstlisting}

\begin{table}[htb!]
\centering
\caption{Documented default parameters for selected models}
\label{tab:llm_parameters}
\small
\begin{tabular}{clccc}
\toprule
Group & Model & Default temperature & Default top-$p$ & Matches $1.0/1.0$\\
\midrule
\multirow{2}{*}{G1}
 & GPT-3.5 Turbo Instruct & 1.0 & 1.0 & \fullmark\\
 & Mixtral 8$\times$22B Instruct & 0.3 & 1.0 & \halfmark\\
\midrule
\multirow{7}{*}{G2}
 & GPT-4o & 1.0 & 1.0 & \fullmark\\
 & Qwen2.5-72B Instruct & 1.0 & 1.0 & \fullmark\\
 & DeepSeek-V3 & 1.0 & 1.0 & \fullmark\\
 & Llama 3.3 70B Instruct & 1.0 & 1.0 & \fullmark\\
 & GLM-4 32B 0414 & 0.75 & 0.90 & \emptymark\\
\midrule
\multirow{9}{*}{G3}
 & DeepSeek-V4-Pro & 1.0 & 1.0 & \fullmark\\
 & Qwen3.6-Max-Preview & 1.0 & 1.0 & \fullmark\\
 & GLM-5.1 & 1.0 & 0.95 & \halfmark\\
 & MiMo-V2.5-Pro & 1.0 & 0.95 & \halfmark\\
 & GPT-5.5 & Unsupported & Unsupported & \unknownmark\\
 & Claude Opus 4.7 & Ignored & Ignored & \unknownmark\\
 & Gemini 3.1 Pro Preview & Not reported & Not reported & \unknownmark\\
 & Kimi K2.6 & 1.0 & 1.0 & \fullmark\\
\bottomrule
\end{tabular}

\begin{minipage}{0.94\linewidth}
\footnotesize
Theses defaulted values are verified against OpenRouter or the model provider's documentation. The final column compares the defaults with the common experimental settings of temperature $=1.0$ and top-$p=1.0$: \fullmark{} indicates that both parameters match, \halfmark{} that one matches, \emptymark{} that neither matches, and \unknownmark{} that the comparison is unavailable. The verification was carried out on 3rd September, 2026. For GPT-5.5 and Claude Opus 4.7, these sampling controls are unsupported or ignored; OpenRouter does not publish defaults for Gemini 3.1 Pro Preview.
\end{minipage}
\end{table}

\newpage

\subsection{Supporting materials for Section~\ref{sec:results}}

\subsubsection{Overall performance}
\label{subsubsec:appendix-overall-performance}

\begin{table}[htb!]
\centering
\caption{Mean performance summaries across generations (\%).}
\label{tab:mean-performance-detector-generator}
\begin{tabular}{ccccc@{\hspace{1.5em}}ccccc}
\toprule
\multicolumn{5}{c}{\textit{Panel A}} &
\multicolumn{5}{c}{\textit{Panel B}}\\
\cmidrule(lr){1-5}\cmidrule(lr){6-10}
& \multicolumn{2}{c}{F1-score} & \multicolumn{2}{c}{Accuracy} &
& \multicolumn{2}{c}{F1-score} & \multicolumn{2}{c}{Accuracy}\\
\cmidrule(lr){2-3}\cmidrule(lr){4-5}\cmidrule(lr){7-8}\cmidrule(lr){9-10}
Detector & Mean & SD & Mean & SD & Generator & Mean & SD & Mean & SD\\
\midrule
GPT-3.5 & 4.8 & 3.9 & 48.0 & 1.1 & GPT-3.5 & 72.7 & 33.3 & 79.2 & 19.9\\
Mixtral & 41.9 & 9.0 & 56.4 & 4.6 & Mixtral & 69.5 & 33.3 & 76.6 & 20.2\\
\midrule
DeepSeek-V3 & 56.1 & 10.8 & 56.2 & 7.1 & DeepSeek-V3 & 70.2 & 33.7 & 77.2 & 20.7\\
GLM-4 & 53.5 & 6.3 & 57.8 & 4.1 & GLM-4 & 69.7 & 33.3 & 76.8 & 20.3\\
GPT-4o & 75.5 & 4.3 & 71.8 & 4.0 & GPT-4o & 71.5 & 33.8 & 78.2 & 20.4\\
Llama-3.3 & 69.6 & 6.7 & 64.1 & 6.0 & Llama-3.3 & \textbf{74.0} & 32.4 & \textbf{80.0} & 19.2\\
Qwen2-72B & 2.5 & 3.8 & 48.5 & 1.1 & Qwen2-72B & 73.1 & 32.9 & 79.3 & 19.7\\
\midrule
Claude-4.7 & 96.4 & 1.2 & 96.3 & 1.1 & Claude-4.7 & 64.1 & 34.6 & 72.8 & 22.4\\
DeepSeek-V4 & 82.1 & 8.0 & 83.9 & 6.2 & DeepSeek-V4 & 66.9 & 32.4 & 74.5 & 20.4\\
Gemini-3.1 & \textbf{99.2} & 1.5 & \textbf{99.2} & 1.4 & Gemini-3.1 & 69.5 & 29.5 & 75.8 & 17.8\\
GLM-5.1 & 82.3 & 4.4 & 84.8 & 3.1 & GLM-5.1 & 71.1 & 28.0 & 76.9 & 17.7\\
GPT-5.5 & 98.9 & 0.2 & 98.8 & 0.2 & GPT-5.5 & 67.7 & 33.7 & 75.3 & 20.6\\
Kimi-K2.6 & 98.9 & 0.5 & 98.9 & 0.5 & Kimi-K2.6 & 67.5 & 30.4 & 74.5 & 19.9\\
MiMo-V2.5 & 83.1 & 7.8 & 84.7 & 6.0 & MiMo-V2.5 & 66.4 & 34.4 & 74.5 & 22.0\\
Qwen3.6 & 98.1 & 1.4 & 98.2 & 1.4 & Qwen3.6 & 68.8 & 33.0 & 76.0 & 20.1\\
\bottomrule
\end{tabular}
\vspace{0.5em}
\begin{minipage}{0.88\linewidth}
\footnotesize
Note: Panel A summarises mean performance of each detector across the 15 generator-specific datasets. Panel B summarises mean performance of all detectors on each generator-specific dataset. SD denotes the sample standard deviation. Bold values indicate the highest mean in each panel.
\end{minipage}
\end{table}

\subsubsection{Self detection and cross detection statistical tests}
\label{subsubsec:appendix-selfcross-statistics}

Before calculating the evaluation metrics and conducting the statistical tests, we assessed the validity of the classification responses. Of the 240,000 responses, 233,428 (97.3\%) contained a valid binary judgement. We considered two types of responses invalid: 1) 1,909 responses (0.8\%) were \emph{missing} because the response field was empty; and 2) 4,663 responses (1.9\%) were \emph{malformed} because their content could not be parsed into either \texttt{human} or \texttt{machine}. Both types were excluded rather than treated as classification errors. All reported performance metrics and statistical comparisons were therefore calculated using the available valid responses.

The main statistical test is for testing whether each detector performs differently on its own generated texts and on texts generated by other models from the same generation. We consider two types of responses as invalid. A response is treated as \emph{missing} when its field is empty. A response is treated as \emph{malformed} when its content cannot be parsed into either \texttt{human} or \texttt{machine}. Both types are excluded rather than treated as classification errors. 

For each detector, self-detection accuracy and F1-score are compared with the unweighted average of its scores of cross-detection from the same generation. We computed $\Delta$Accuracy and $\Delta$F1 as difference between the self-detection score and the corresponding cross-detection score. Hence, a positive difference indicates better self-detection performance, while a negative difference indicates better cross-detection performance.

We used a prompt-clustered bootstrap with 10,000 iterations to estimate 95\% percentile confidence intervals and two-sided null-centred bootstrap tests. The smallest reportable $p$-value is therefore $p<0.0001$. Because a separate hypothesis is tested for each of the 15 detectors, detector-level p-values are adjusted using Holm's procedure to control the family-wise error rate. For model-level tests, a difference (i.e., $\Delta$Accuracy or $\Delta$F1) is considered statistically significant when its Holm-adjusted $p$-value is under 0.05. For generation-level tests, the difference between self- and cross-detection are obtained by averaging the detector-level differences within each generation for every bootstrap iteration.

In addition to above, we also aggregated the detector-level results within each generation as reported in Table~\ref{tab:within-generation-level-mean}. G2 detectors perform significantly worse in self-detection than in matched same-generation cross-detection, with an accuracy difference of $-1.65$ percentage points (95\% CI $[-2.14,-1.15]$). No generation-level difference has been found for G1 or G3 as their confidence intervals include zero.

Furthermore, two more sensitivity tests are conducted. For the first analysis, we compared each detector's self-detection performance with its performance on texts generated by all 14 other models. We assigned each of the 14 non-self generator models equal weight to compute cross-detection performance. The self-detection accuracy is 0.85 percentage points lower than cross-detection accuracy (95\% CI $[-1.12,-0.58]$) on average. For the second analysis, as the three generations contain different numbers of models, we assigned equal total weights to G1, G2, and G3 to calculate cross-detection performance. Under this generation-balanced comparison, self-detection accuracy is 1.42 percentage points lower (95\% CI $[-1.69,-1.15]$) on average. 
Despite both differences are statistically significant ($p<0.0001$), their magnitudes are small. Overall, as shown in Table~\ref{tab:allgen-model-weighted} and Table~\ref{tab:allgen-generation-balanced} in Appendix~\ref{subsubsec:appendix-selfcross-statistics}, six detectors show significant differences under both analyses, while GPT-4o is significant only in the first analysis. In summary, the results confirm that the average differences between self- and cross-detection are small, while their direction and magnitude vary across individual detectors.

\begin{table}[htb!]
\centering
\caption{Generation-level within-generation self- and cross-detection results.}
\label{tab:within-generation-level-mean}
\adjustbox{max width=\columnwidth}{
\begin{tabular}{ccccccc}
\toprule
Metric & Generation & Self (\%) & Matched cross (\%) & $\Delta$ (pp) & 95\% CI (pp) & $p_{\mathrm{Holm}}$ \\
\midrule
\multirow{3}{*}{Accuracy}
 & G1 & 51.1 & 51.9 & $-0.83$ & $[-1.79,+0.16]$ & 0.1876 \\
 & G2 & 60.6 & 62.3 & $\mathbf{-1.65}$ & $\mathbf{[-2.14,-1.15]}$ & $\mathbf{0.0003}$ \\
 & G3 & 91.3 & 91.5 & $-0.21$ & $[-0.59,+0.17]$ & 0.2696 \\
\midrule
\multirow{3}{*}{F1-score}
 & G1 & 20.8 & 22.4 & $-1.62$ & $[-3.73,+0.58]$ & 0.2860 \\
 & G2 & 52.3 & 54.8 & $\mathbf{-2.45}$ & $\mathbf{[-3.20,-1.71]}$ & $\mathbf{0.0003}$ \\
 & G3 & 90.1 & 90.4 & $-0.28$ & $[-0.80,+0.25]$ & 0.3017 \\
\bottomrule
\end{tabular}
}
\begin{minipage}{0.92\columnwidth}
\footnotesize
Note: Self and matched-cross scores are percentages; differences and confidence intervals are percentage points. The two-sided bootstrap p-values are Holm-adjusted across G1, G2, and G3 separately for each metric; significant results at $\alpha=0.05$ are shown in bold.
\end{minipage}
\end{table}

\begin{table}[htb!]
\centering
\caption{Detector-level model-weighted all-generation sensitivity results.}
\label{tab:allgen-model-weighted}
\scriptsize
\begin{tabular}{c cccc ccc}
\toprule
& & \multicolumn{3}{c}{Accuracy} & \multicolumn{3}{c}{F1-score}\\
\cmidrule(lr){3-5}\cmidrule(lr){6-8}
Generation & Detector & $\Delta$ (pp) & 95\% CI (pp) & $p_{\mathrm{Holm}}$ & $\Delta$ (pp) & 95\% CI (pp) & $p_{\mathrm{Holm}}$\\
\midrule
G1 & GPT-3.5 Turbo Instruct & $-0.56$ & $[-0.96,-0.12]$ & 0.0632 & $-1.96$ & $[-3.36,-0.43]$ & 0.0686\\
G1 & Mixtral 8x22B & $-1.84$ & $[-3.20,-0.49]$ & 0.0632 & $-3.45$ & $[-6.45,-0.51]$ & 0.1338\\
\midrule
G2 & DeepSeek-V3 & \textbf{$-3.55$} & $[-5.06,-1.96]$ & \textbf{0.0015} & \textbf{$-4.44$} & $[-6.76,-2.07]$ & \textbf{0.0018}\\
G2 & GLM-4-32B & $+0.11$ & $[-1.33,+1.52]$ & 1.0000 & $+0.47$ & $[-1.87,+2.71]$ & 1.0000\\
G2 & GPT-4o & \textbf{$+1.73$} & $[+0.93,+2.51]$ & \textbf{0.0015} & \textbf{$+1.98$} & $[+1.14,+2.79]$ & \textbf{0.0015}\\
G2 & Llama 3.3 70B & \textbf{$+7.12$} & $[+6.41,+7.83]$ & \textbf{0.0015} & \textbf{$+7.67$} & $[+6.90,+8.46]$ & \textbf{0.0015}\\
G2 & Qwen2-72B & $-0.25$ & $[-0.50,+0.04]$ & 0.3470 & $-0.84$ & $[-1.78,+0.24]$ & 0.4092\\
\midrule
G3 & Claude Opus 4.7 & $+0.29$ & $[+0.06,+0.49]$ & 0.0632 & $+0.31$ & $[+0.07,+0.51]$ & 0.0608\\
G3 & DeepSeek-V4-Pro & \textbf{$-5.16$} & $[-6.70,-3.65]$ & \textbf{0.0015} & \textbf{$-6.34$} & $[-8.52,-4.26]$ & \textbf{0.0015}\\
G3 & Gemini 3.1 Pro Preview & \textbf{$-5.34$} & $[-6.35,-4.37]$ & \textbf{0.0015} & \textbf{$-5.69$} & $[-6.83,-4.61]$ & \textbf{0.0015}\\
G3 & GLM-5.1 & \textbf{$-7.14$} & $[-8.95,-5.27]$ & \textbf{0.0015} & \textbf{$-10.32$} & $[-13.21,-7.41]$ & \textbf{0.0015}\\
G3 & GPT-5.5 & $+0.01$ & $[-0.08,+0.08]$ & 1.0000 & $+0.01$ & $[-0.08,+0.08]$ & 1.0000\\
G3 & Kimi K2.6 & $+0.03$ & $[-0.36,+0.38]$ & 1.0000 & $+0.03$ & $[-0.36,+0.39]$ & 1.0000\\
G3 & MiMo V2.5 Pro & $+0.93$ & $[-0.25,+2.07]$ & 0.4840 & $+1.63$ & $[+0.17,+3.04]$ & 0.1338\\
G3 & Qwen3.6 Max Preview & \textbf{$+0.89$} & $[+0.65,+1.11]$ & \textbf{0.0015} & \textbf{$+0.93$} & $[+0.68,+1.15]$ & \textbf{0.0015}\\
\bottomrule
\end{tabular}
\begin{minipage}{0.96\linewidth}
\footnotesize
Note: Differences and confidence intervals are reported in percentage points. The cross-detection baseline includes all 14 non-self generators, each receiving equal weight. $\Delta$ is self-detection minus cross-detection. Confidence intervals are obtained from 10,000 prompt-clustered bootstrap iterations. Holm-adjusted $p$-values control for 15 detector-level tests; significant results at $\alpha=0.05$ are shown in bold.
\end{minipage}
\end{table}

\begin{table}[htb!]
\centering
\caption{Detector-level generation-balanced all-generation sensitivity results.}
\label{tab:allgen-generation-balanced}
\small
\setlength{\tabcolsep}{3.2pt}
\begin{tabular}{c cccc ccc}
\toprule
& & \multicolumn{3}{c}{Accuracy} & \multicolumn{3}{c}{F1-score}\\
\cmidrule(lr){3-5}\cmidrule(lr){6-8}
Generation & Detector & $\Delta$ (pp) & 95\% CI (pp) & $p_{\mathrm{Holm}}$ & $\Delta$ (pp) & 95\% CI (pp) & $p_{\mathrm{Holm}}$\\
\midrule
G1 & GPT-3.5 Turbo Instruct & $-0.28$ & $[-0.67,+0.17]$ & 1.0000 & $-0.93$ & $[-2.34,+0.59]$ & 1.0000\\
G1 & Mixtral 8x22B & $-1.81$ & $[-3.21,-0.39]$ & 0.0999 & $-3.60$ & $[-6.68,-0.58]$ & 0.1863\\
\midrule
G2 & DeepSeek-V3 & \textbf{$-5.29$} & $[-6.80,-3.69]$ & \textbf{0.0015} & \textbf{$-7.05$} & $[-9.37,-4.68]$ & \textbf{0.0015}\\
G2 & GLM-4-32B & $-0.49$ & $[-1.96,+0.96]$ & 1.0000 & $-0.57$ & $[-2.92,+1.72]$ & 1.0000\\
G2 & GPT-4o & $+0.67$ & $[-0.13,+1.44]$ & 0.7551 & $+0.83$ & $[-0.02,+1.62]$ & 0.3856\\
G2 & Llama 3.3 70B & \textbf{$+5.42$} & $[+4.77,+6.10]$ & \textbf{0.0015} & \textbf{$+5.78$} & $[+5.08,+6.50]$ & \textbf{0.0015}\\
G2 & Qwen2-72B & $-0.11$ & $[-0.35,+0.17]$ & 1.0000 & $-0.33$ & $[-1.25,+0.71]$ & 1.0000\\
\midrule
G3 & Claude Opus 4.7 & $+0.11$ & $[-0.12,+0.31]$ & 1.0000 & $+0.12$ & $[-0.11,+0.32]$ & 1.0000\\
G3 & DeepSeek-V4-Pro & \textbf{$-6.72$} & $[-8.27,-5.23]$ & \textbf{0.0015} & \textbf{$-8.32$} & $[-10.51,-6.26]$ & \textbf{0.0015}\\
G3 & Gemini 3.1 Pro Preview & \textbf{$-5.43$} & $[-6.44,-4.46]$ & \textbf{0.0015} & \textbf{$-5.78$} & $[-6.93,-4.70]$ & \textbf{0.0015}\\
G3 & GLM-5.1 & \textbf{$-7.24$} & $[-9.06,-5.35]$ & \textbf{0.0015} & \textbf{$-10.46$} & $[-13.37,-7.55]$ & \textbf{0.0015}\\
G3 & GPT-5.5 & $-0.03$ & $[-0.12,+0.05]$ & 1.0000 & $-0.03$ & $[-0.12,+0.05]$ & 1.0000\\
G3 & Kimi K2.6 & $-0.09$ & $[-0.47,+0.25]$ & 1.0000 & $-0.08$ & $[-0.47,+0.26]$ & 1.0000\\
G3 & MiMo V2.5 Pro & $-0.67$ & $[-1.84,+0.48]$ & 1.0000 & $-0.43$ & $[-1.86,+0.98]$ & 1.0000\\
G3 & Qwen3.6 Max Preview & \textbf{$+0.58$} & $[+0.35,+0.77]$ & \textbf{0.0015} & \textbf{$+0.60$} & $[+0.37,+0.80]$ & \textbf{0.0015}\\
\bottomrule
\end{tabular}
\begin{minipage}{0.96\linewidth}
\footnotesize
Note: Differences and confidence intervals are reported in percentage points. The cross-detection baseline includes all 14 non-self generators. G1, G2, and G3 receive equal total weight, which is divided equally among the models within each generation. $\Delta$ is self-detection minus cross-detection. Confidence intervals are obtained from 10,000 prompt-clustered bootstrap iterations. Holm-adjusted $p$-values control for 15 detector-level tests; significant results at $\alpha=0.05$ are shown in bold.
\end{minipage}
\end{table}

\newpage
\subsubsection{False-positive and false-negative patterns}
\label{subsubsec:appendix-fp-fn-patterns}

\begin{figure*}[htb!]
\centering
\includegraphics[width=\linewidth]{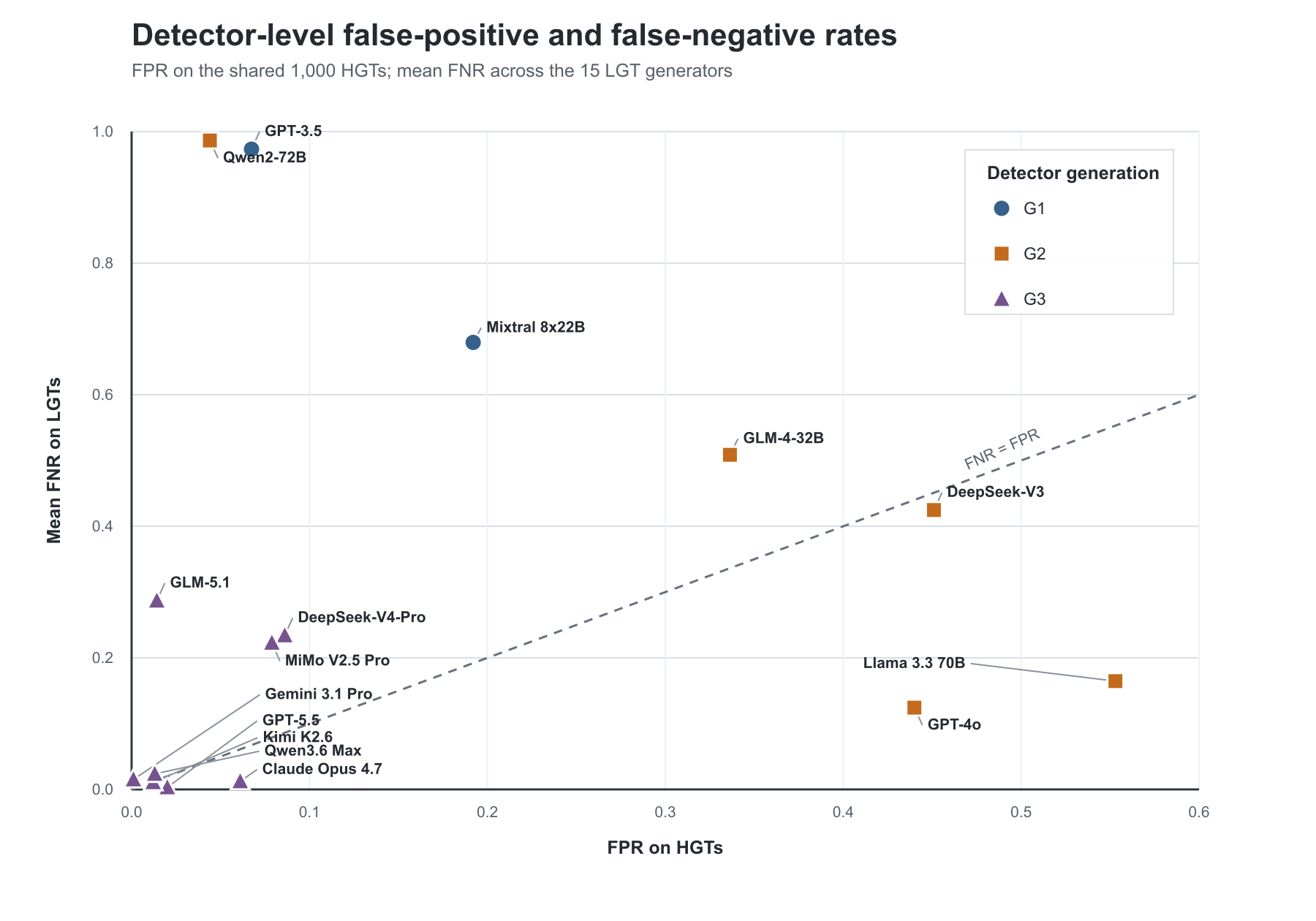}
\caption{Detector-level error profiles for the 15 selected models.}
\label{fig:detector-error-profiles}
\end{figure*}

Figure~\ref{fig:detector-error-profiles} illustrates the detector-level error profiles for all 15 selected LLM models. Each point in the figures presents a detector's FPR on the shared HGT corpus and its mean FNR across texts produced by all 15 generators across generations. Detectors in different generations are color coded as indicated in the legend. The dashed line indicates equal FNR and FPR. Points above the dashed line make proportionally more false-negative errors, whereas points below it make proportionally more false-positive errors. As shown in Figure~\ref{fig:detector-error-profiles}, individual detectors within the same generation can exhibit different error patterns. On top-left corner of the figure, GPT-3.5 Turbo Instruct and Qwen2-72B have high mean FNRs but low FPRs, indicating a strong tendency to assign the human label. In contrast, on the right-bottom region of the figure, GPT-4o and Llama 3.3 70B reduce FNRs but have comparatively high FPRs. Most G3 detectors are concentrated in the lower-left region, indicating low rates for both FNRs and FPRs. Nevertheless, GLM-5.1, DeepSeek-V4-Pro, and MiMo V2.5 Pro are more error-prone compared with the rest of G3 detectors. 

\subsubsection{Semantic error pattern analysis}
\label{subsubsec:appendix-semantic-error}

It is impractical to perform a full manual analysis of all explanations, to facilitate the analyses, we selected two generators and two detectors from each generation. The generators selected are GPT-3.5 Turbo Instruct and Mixtral 8x22B (G1), GPT-4o and DeepSeek-V3 (G2), and GPT-5.5 and MiMo V2.5 Pro (G3). The detectors selected are the same G1 and G2 models, together with GPT-5.5 and DeepSeek-V4-Pro for G3. 

\begin{table}[htb!]
\centering
\footnotesize
\caption{Overview of the sampled semantic-analysis dataset.}
\label{tab:semantic-sample-summary}
\begin{tabular}{ccccc}
\toprule
\multicolumn{5}{c}{\textit{Panel A: Detector agreement by text type}}\\
\midrule
Text type & Texts & Disagreement & Pairwise comparisons & Label agreement \%\\
\midrule
HGT & 20 & 8 & 300 & 83.7\\
LGT & 120 & 120 & 1,800 & 48.4\\
\midrule
\multicolumn{5}{c}{\textit{Panel B: Classification outcomes by detector}}\\
\midrule
Detector & TP & TN & FP & FN\\
\midrule
GPT-3.5 Turbo Instruct (G1) & 0 & 20 & 0 & 120\\
Mixtral 8x22B (G1) & 30 & 19 & 1 & 90\\
GPT-4o (G2) & 93 & 15 & 5 & 27\\
DeepSeek-V3 (G2) & 70 & 17 & 3 & 50\\
GPT-5.5 (G3) & 120 & 20 & 0 & 0\\
DeepSeek-V4-Pro (G3) & 93 & 18 & 2 & 27\\
\bottomrule
\end{tabular}
\end{table}

We randomly sampled 20 text generation prompts. For every sampled generation prompt, we retained its associated HGT and all six LGTs produced by the selected generators. Direction 1 therefore contains 140 unique texts: 20 HGTs and 120 LGTs ($20$ prompts $\times$ $6$ generators). Each text is judged by six detectors, producing 120 explanations for the HGTs and 720 explanations for the LGTs (i.e, 840 explanations in total). Direction 2 reorganises the 720 LGT explanations into 120 detector--text-generation-prompt comparison groups ($6$ detectors $\times$ $20$ prompts), with each group comparing the six generator outputs.

Table~\ref{tab:semantic-sample-summary} provides an overview of this sampled dataset. Panel A reports whether the six detectors agree on the label assigned to each text, while Panel B reports the resulting classification outcomes for each detector. All 120 LGTs produce disagreement among the six detectors, compared with eight of the 20 HGTs. Across the 840 judgements, TP and FN sum to 720 judgements on LGTs, whereas TN and FP sum to 120 judgements on HGTs.

\subsubsection{LLM-assisted semantic error pattern analysis}
\label{subsubsec:appendix-llm}

The LLM-assisted analysis uses separate prompts for Direction 1, Direction 2, and the synthesis of the two outputs. 

\begin{lstlisting}[basicstyle=\ttfamily\scriptsize,breaklines=true,frame=lines,caption={Direction 1 review prompt},label={lst:direction-a-review-prompt}]
Task: Semantic pattern discovery in detector explanations (Direction 1)

You are reviewing explanations from an LLM-generated-text detection study. Each record holds one exact text constant and presents judgements from six detectors: two from G1, two from G2, and two from G3. The records include HGTs and LGTs from two representative generators per generation.

Analyse the explanations semantically rather than merely counting keywords. Distinguish an accurate observation about the text from a valid inference about the classification label. A grounded explanation can still be logically unreliable.

Report:
1. reasoning strategies used by each detector and detector generation;
2. whether explanations cite concrete textual evidence or use generic claims;
3. whether detectors that agree on a label also agree on the reason;
4. recurring rationale patterns preceding false positives and false negatives;
5. whether the same textual cue is interpreted in opposite directions;
6. confidence framing and whether confidence appears to track correctness; and
7. similarities and differences across detector generations.

Use only the supplied records. Cite evidence as source + text_id + detector. Do not invent frequencies. Give exact counts only when they can be directly tallied from the records; otherwise use qualitative terms such as recurring or occasional. Treat generation-wide interpretations as exploratory because only two detectors represent each generation.

Output:
- an executive summary;
- a table of named semantic patterns, affected detectors/outcomes, and example record references;
- detector-level and generation-level findings;
- contradictory or inverted cue examples;
- limitations; 

Data follows.
\end{lstlisting}

\begin{lstlisting}[basicstyle=\ttfamily\scriptsize,breaklines=true,frame=lines,caption={Direction 2 review prompt},label={lst:direction-b-review-prompt}]
Task: Semantic pattern discovery in detector explanations (Direction 2)

You are reviewing explanations from an LLM-generated-text detection study. Each group holds the detector and text generation prompt constant while varying the LGT source across six generators: two from G1, two from G2, and two from G3. All texts in this bundle are machine-generated. This view tests whether the same detector changes its label or reasoning for different generator outputs written in response to the same text generation prompt.

Analyse the explanations semantically rather than merely counting keywords. Distinguish genuine sensitivity to textual evidence from a detector that appears consistent only because it repeatedly returns the same label.

Report:
1. detector-specific stability and changes across generator outputs;
2. generator or generator-generation properties associated with changed labels;
3. whether a stable label is accompanied by stable reasoning;
4. recurring false-negative rationales, especially on G3-generated texts;
5. cases where the same detector interprets similar cues differently; and
6. implications for comparing self- and cross-detection explanations.

Use only the supplied records. Cite evidence as detector + text_id + source. Do not invent frequencies. Give exact counts only when they can be directly tallied from the records; otherwise use qualitative terms. Treat generation-wide interpretations as exploratory because only two generators represent each generation.

Output:
- an executive summary;
- a table of named semantic patterns, affected detectors/sources, and example record references;
- detector-level and generator-level findings;
- limitations; 

Data follows.
\end{lstlisting}

Direction 1 examines whether detectors reach the same judgement and use the same reasoning for the same text. For each text, the LLM first identifies the textual properties all explanations and then compares how each detector link those properties to a ``human'' or ``machine'' label. The LLM identifies five recurring forms of disagreement via this process. As reported in Table~\ref{tab:direction-one-semantic-examples}, detectors can notice same characteristics of the writing but interpret them differently to make different decisions. For example, detailed and coherent writing may be treated as evidence of human expertise by one detector, but as evidence of formulaic LLM generation by another. Similar observations of disagreements arise for personal voice, formal style, and textual artefact.

Direction 2 examines whether a detector applies consistent reasoning to outputs generated by different generators with the same text generation prompt. We observed that Mixtral 8x22B , GPT-4o, DeepSeek-V3, and DeepSeek-V4-Pro change their labels on text generated by different generators, and Table~\ref{tab:direction-two-semantic-examples} shows examples of how GPT-4o, DeepSeek-V3 and DeepSeek-V4-Pro as detectors change their judgements on text generated by the same prompt across three generations. For the `academic-pressure' prompt, GPT-4o treats empathy and conversational language as human evidence in four outputs, but classifies two more regularly structured dialogues as machine-generated. For the `unreliable-narration' prompt, DeepSeek-V3 classifies the uniformly structured GPT-3.5 output as machine-generated but treats the Mixtral output as human-written because it integrates literary examples and appears more nuanced. For the `climate-architecture' prompt, DeepSeek-V4-Pro classifies all G1 and G2 outputs as machine-generated but both G3 outputs as human-written, describing the latter as more natural and less formulaic. These comparisons show that the judgement and explanation can change with the generated text even when the detector and text generation prompt (i.e., same topic under the same rules) remain unchanged.

\begin{table}[htb!]
\centering
\scriptsize
\setlength{\tabcolsep}{3pt}
\caption{Direction 1 examples: identical texts judged by different detectors.}
\label{tab:direction-one-semantic-examples}
\begin{tabular}
{p{0.08\linewidth}p{0.05\linewidth}p{0.20\linewidth}p{0.22\linewidth}p{0.22\linewidth}p{0.12\linewidth}}
\toprule
Generator & Prompt ID & Shared text excerpt & Detector 1: judgement and explanation & Detector 2: judgement and explanation & Observed difference \\
\midrule
GPT-3.5 & 40 & ``Unreliable narration is a literary technique where the narrator's credibility is called into question.'' & GPT-3.5: H (FN). Detailed literary analysis is described as beyond an LLM's capability. & GPT-5.5: M (TP). Generic organisation, repeated claims, standard transitions, and the lack of concrete examples support the machine label. & The detectors draw opposite conclusions about the depth of the same text. \\
Human & 43 & Factual history of Brentford Football Club. & GPT-3.5: H (TN). Specific records, statistics, and recent updates are treated as human evidence. & GPT-5.5: H (TN). Article cross-references and an incomplete phrase are treated as signs of a copied human-edited source. & The detectors agree on the label but rely on different reasons. \\
GPT-3.5 & 113 & ``Medical malpractice occurs when a healthcare provider fails to follow the accepted standard of care.'' & DeepSeek-V3: H (FN). Specific examples, coherence, and a detailed approach are treated as human evidence. & DeepSeek-V4-Pro: M (TP). Predictable transitions, generic coverage, and uniform tone are treated as LGT cues. & The same organisation and detail support opposite labels and reasons. \\
Mixtral & 146 & ``I've always found that the true essence of a city lies beyond its well-trodden tourist paths.'' & Mixtral: H (FN). Sensory details and personal recollection are interpreted as lived experience. & GPT-5.5: M (TP). Familiar travel phrases and a decorative anecdote are interpreted as generated personalisation. & The same personal voice is treated as genuine or simulated. \\
Human & 425 & ``Indian Mary Park is part of the Josephine County Parks system.'' & DeepSeek-V4-Pro: M (FP). Missing distances are interpreted as model generation errors. & GPT-5.5: H (TN). The same omissions are interpreted as extraction or formatting artefacts. & The same textual defect supports opposite provenance inferences. \\
\bottomrule
\end{tabular}
\begin{minipage}{0.96\linewidth}
\scriptsize
Note: H and M denote predicted human and machine labels, respectively. Within each row, both detectors judge the same complete text; the third column presents a short excerpt from that shared text. TP, TN, FP, and FN denote the corresponding classification outcomes.
\end{minipage}
\end{table}

\begin{table}[htb!]
\centering
\scriptsize
\setlength{\tabcolsep}{2pt}
\caption{Direction 2 examples: one detector judging outputs generated from the same prompt.}
\label{tab:direction-two-semantic-examples}
\begin{tabular}
{p{0.1\linewidth}p{0.1\linewidth}p{0.20\linewidth}p{0.19\linewidth}p{0.19\linewidth}p{0.13\linewidth}}
\toprule
Detector & Prompt ID and topic & Generators and output excerpts & Output 1: judgement and explanation & Output 2: judgement and explanation & Observed difference \\
\midrule
GPT-4o & 43: academic pressure & GPT-3.5: ``Hey there! How have you been dealing with all the academic pressure lately?'' \newline GPT-4o: ``Alice: Hey Ben, you've been looking a bit stressed lately.'' & GPT-3.5 output: H (FN). Empathy, emotional support, and conversational markers are treated as human evidence. & GPT-4o output: M (TP). Sequential advice and a deliberate narrative arc are treated as machine evidence. & Different dialogues lead the same detector to opposite labels and reasons. \\
DeepSeek-V3 & 40: unreliable narration & GPT-3.5: ``Unreliable narration is a literary technique where the narrator's credibility is called into question.'' \newline Mixtral: ``Unreliable narration is a narrative technique where the credibility of a narrator is compromised.'' & GPT-3.5 output: M (TP). Uniform structure, systematic coverage, and repetition support the machine label. & Mixtral output: H (FN). Nuanced analysis and integrated literary examples support the human label. & The judgement changes with the depth and form of the generated text. \\
DeepSeek-V4-Pro & 34: climate-responsive architecture & GPT-3.5: ``Climate change is a global issue that affects every aspect of our lives, including the design of our buildings.'' \newline GPT-5.5: ``Future architectural designs are likely to become more climate-responsive, flexible, and resource-efficient.'' & GPT-3.5 output: M (TP). Formulaic organisation, generic language, and predictable progression support the machine label. & GPT-5.5 output: H (FN). Progressive examples and the absence of awkward phrasing support the human label. & Outputs from different generator generations lead to opposite judgements. \\
\bottomrule
\end{tabular}
\begin{minipage}{0.96\linewidth}
\scriptsize
Note: The detector and text generation prompt are fixed within each row, while the generator and resulting text change. Output 1 and Output 2 correspond, in order, to the two generators listed in the third column. H and M denote predicted human and machine labels; because all evaluated texts are LGTs, H is an FN and M is a TP.
\end{minipage}
\end{table}

\newpage
\subsubsection{Cue-based analysis}
\label{subsubsec:appendix-cue}

\begin{table}[htb!]
\centering
\scriptsize
\caption{Explanation cue categories and their links to prior literature.}
\label{tab:semantic-cue-literature}
\begin{tabular}{p{0.22\linewidth}p{0.2\linewidth}p{0.4\linewidth}}
\toprule
Cue category & Interpretation in detector explanations & Supporting literature\\
\midrule
Structure and formatting & Organisation, paragraph structure, formatting, or ordering of ideas & Previous studies found that human evaluators use formatting, textual structure, and sentence organisation to make a judgment~\citep{Clark-E2021, Russell-J2025}.\\
Fluency and polish & Grammar, spelling, fluency, formality, smoothness, or tonal consistency & Grammar, fluency, formality, and clarity are commonly used as detection cues, although they may support either a human or machine judgement~\citep{Clark-E2021, Mitrovic-S2023, Russell-J2025}.\\
Generic or abstract language & Broad, generic, shallow, or insufficiently original content & LGTs are often perceived as focusing on general concepts and lacking detail or originality~\citep{Mitrovic-S2023, Russell-J2025}.\\
Hedging and uncertainty & Qualified claims and expressions of uncertainty or epistemic caution & Human evaluators sometimes use expressions of uncertainty to make a judgment, although uncertainty is not a consistently reliable detection cue~\citep{Clark-E2021}.\\
Personal voice & Personal pronouns, feelings, anecdotes, lived experience, or individual perspective & Personal expression and emotional language are often associated with human writing and may cause LGTs to be misclassified as human-written~\citep{Clark-E2021, Mitrovic-S2023, Russell-J2025}.\\
Coherence and transitions & Logical flow, consistency, repetition, cohesion, or formulaic connections & Previous studies found that evaluators frequently consider coherence, repetition, and sentence structure when making a judgment~\citep{Clark-E2021, Russell-J2025}.\\
Specificity and evidence & Concrete details, examples, citations, data, or other supporting evidence & Evaluators often use detail and specificity to make a judgment, but these cues can also lead to incorrect decisions~\citep{Clark-E2021, Mitrovic-S2023, Russell-J2025}.\\
Factuality and hallucination & Factual errors, contradictions, implausible claims, or unsupported content & Factual errors and contradictions may be used as signs of machine generation, while hallucination is a recognised problem in generated text~\citep{Ippolito-D2020, Clark-E2021, Ji-Z2023}.\\
\bottomrule
\end{tabular}
\end{table}

\begin{table}[htb!]
\centering
\scriptsize
\caption{Keywords and phrases used to identify explanation cues.}
\label{tab:semantic-cue-keywords}
\begin{tabular}{p{0.22\linewidth}p{0.70\linewidth}}
\toprule
Cue category & Predefined keywords and phrases\\
\midrule
Structure and formatting & \texttt{structured}; \texttt{format}; \texttt{paragraph}; \texttt{introduction}; \texttt{conclusion}; \texttt{essay}; \texttt{organization}\\
Fluency and polish & \texttt{polished}; \texttt{fluent}; \texttt{smooth}; \texttt{grammatically}; \texttt{well-formed}; \texttt{well formed}; \texttt{even tone}\\
Generic or abstract language & \texttt{generic}; \texttt{abstract}; \texttt{high level}; \texttt{encyclopedic}; \texttt{surface-level}; \texttt{surface level}; \texttt{broad}\\
Hedging and uncertainty & \texttt{hedge}; \texttt{hedged}; \texttt{hedging}; \texttt{in most cases}; \texttt{tends to}; \texttt{may}; \texttt{might}; \texttt{depending on}\\
Personal voice & \texttt{personal voice}; \texttt{idiosyncr*}; \texttt{anecdote}; \texttt{lived experience}; \texttt{individual perspective}\\
Coherence and transitions & \texttt{transition}; \texttt{coheren*}; \texttt{flow}; \texttt{parallel}; \texttt{balanced}; \texttt{systematic*}\\
Specificity and evidence & \texttt{specific}; \texttt{citation}; \texttt{example}; \texttt{concrete}; \texttt{detail}; \texttt{data}; \texttt{factual}\\
Factuality and hallucination & \texttt{hallucination}; \texttt{factual error}; \texttt{inaccur*}; \texttt{wrong}; \texttt{confusion}; \texttt{contradict*}\\
\bottomrule
\end{tabular}
\begin{minipage}{0.94\linewidth}
\footnotesize
Note: Matching is case-insensitive and requires complete words unless an asterisk is shown. An asterisk denotes any continuation of the word stem; for example, \texttt{coheren*} matches \texttt{coherent} and \texttt{coherence}. Hyphenated and unhyphenated forms are both matched where both forms are listed. Each cue category is counted at most once within an explanation.
\end{minipage}
\end{table}

To help identify the eight cue categories reported in Table~\ref{tab:semantic-cue-literature} among LLM-generated explanations, we predefined keywords and phrases as reported in Table~\ref{tab:semantic-cue-keywords}. We first converted each explanation to lowercase, then search these terms in each explanation. A category is recorded when at least one of its associated terms are identified in the explanation. One explanation may contain multiple categories, thereby a match indicates that the detector refers to a cue in its explanation but this cannot be considered as direct support of the corresponding binary classification decision.

We then compared cue frequencies between true-positive and false-negative explanations for each detector. As reported in Panel B of Table~\ref{tab:semantic-sample-summary}, GPT-3.5 labels all 120 sampled LGTs as human and therefore has no true positives, whereas GPT-5.5 correctly labels all 120 LGTs and has no false negatives. Within-detector comparisons are therefore available only for Mixtral, GPT-4o, DeepSeek-V3, and DeepSeek-V4-Pro, which have observations in both outcome groups.

Furthermore, we observed that four cue categories are largely consistent with the recurring findings of LLM-assisted analysis results. \emph{Structure and formatting} captures references to regular or formulaic organisation, while \emph{generic or abstract language} captures references to broad and insufficiently distinctive coverage. The LLM-assisted analysis found that detectors repeatedly use these two cue categories to support machine judgements. In addition, \emph{specificity and evidence} corresponds to the finding from LLM-assisted analysis that suggests detailed examples, technical knowledge, or apparent expertise can lead detectors to incorrectly assign the human label. \emph{Coherence and transitions} is the cue-category corresponding to cue-inversion pattern observed in the LLM-assisted analysis, where the same clear and logically organised structure can lead different detectors to assign opposite labels. For instance, some detectors treat such patterns as evidence of thoughtful human writing, whereas others treat it as a formulaic pattern typical of LLM-generated text. Since both LLM-assisted analysis and cue-based analysis converge on these four categories, we computed how frequently each cue is mentioned in true-positive and false-negative explanations, as reported in Table~\ref{tab:cue-analysis-results}.

\emph{Structure and formatting} are considerably more frequent in true-positive than false-negative explanations for every detector. For example, this cue appears in 91.4\% of GPT-4o's true positives but only 33.3\% of its false negatives. In one true-positive example, DeepSeek-V4-Pro describes a explanation generated by G1 model (i.e., GPT-3.5) as following a ``highly structured and formulaic pattern''. \emph{Generic or abstract language} follows a similar pattern. These results support the LLM-assisted finding that detectors commonly associate regular organisation and generic coverage with machine generation. \emph{Specificity and evidence} shows the opposite tendency. For Mixtral, GPT-4o, DeepSeek-V3, and DeepSeek-v4-Pro, explanations that have specific details and evidence are more frequently judged as false-negative than true-positive. For instance, DeepSeek-V3 refers to ``specific examples'' and a ``detailed and structured approach'' as reasoning evidence to incorrectly classify an explanation generated by G1 model (i.e., GPT-3.5) as human-written. This is consistent with the LLM-assisted analysis examples in Table~\ref{tab:direction-one-semantic-examples}, where technical details, literary examples, and sensory descriptions are interpreted as evidence of human authorship. \emph{Coherence and transitions}, however, appears frequently in both correct and incorrect judgement. For example, it appears in 83.3\% of Mixtral's true-positive explanations and 85.6\% of its false-negative explanations. This supports the cue-inversion finding, where detectors frequently mention coherence in their reasoning, but they may interpret it as either naturally organised human writing or predictable LLM behaviour. Overall, the cue-based results provide convergent evidence for several patterns identified by the LLM-assisted analysis, while also showing that the same cue does not consistently lead to the same decision of human/LLM authorship.

\begin{table}[htb!]
\centering
\small
\caption{Cue frequencies in correct and incorrect LGT judgements.}
\label{tab:cue-analysis-results}
\begin{tabular}{cc cccc}
\toprule
Detector & Label ($n$) & SF & GAL & CT & SE\\
\midrule
GPT-3.5 Turbo Instruct & FN (120) & 10.0 & 2.5 & 20.0 & 41.7\\
\midrule
Mixtral 8x22B & TP (30) & 93.3 & 73.3 & 83.3 & 40.0\\
 & FN (90) & 61.1 & 46.7 & 85.6 & 65.6\\
\midrule
GPT-4o & TP (93) & 91.4 & 20.4 & 77.4 & 34.4\\
 & FN (27) & 33.3 & 11.1 & 66.7 & 48.1\\
\midrule
DeepSeek-V3 & TP (70) & 90.0 & 57.1 & 78.6 & 32.9\\
 & FN (50) & 28.0 & 40.0 & 80.0 & 42.0\\
\midrule
GPT-5.5 & TP (120) & 83.3 & 98.3 & 76.7 & 91.7\\
\midrule
DeepSeek-V4-Pro & TP (93) & 93.5 & 87.1 & 63.4 & 62.4\\
 & FN (27) & 40.7 & 63.0 & 66.7 & 77.8\\
\bottomrule
\end{tabular}
\begin{minipage}{0.94\linewidth}
\footnotesize
Note: SF = Structure and Formatting; GAL = Generic or Abstract Language; CT = Coherence and Transitions; SE = Specificity and Evidence.
Values are percentages of explanations containing at least one predefined term from the corresponding cue category. An explanation can contain multiple categories. GPT-3.5 has no true positives and GPT-5.5 has no false negatives in this review sample, so within-detector contrasts are unavailable for these models.
\end{minipage}
\end{table}

\subsection{Supporting materials for Section~\ref{sec:discussions-limitation-future-work}}
\label{subsec:appendix-discussion}

Table~\ref{tab:watermark-llms} presents our exploratory review on public documentation related to watermarking from different LLM vendors. These public documentations were reviewed on 14 September 2026. ``No public model- and endpoint-specific claim identified'' does not demonstrate that a watermark is absent, instead, it means that we found no documentation confirming its use for the endpoint through which the experimental text was generated. Google limits its public deployment claim to the Gemini app and web experience~\cite{Google-G2024}. Anthropic describes a future-model rollout and a transition period for older models~\cite{Anthropic-A2026b}. OpenAI documents deployed provenance signals for images and audio while describing text provenance as ongoing work~\cite{OpenAI-O2026}. $\Delta$Accuracy is self-detection accuracy minus matched same-generation cross-detection accuracy. Positive values represent self-detection advantage while negative values represent self-detection disadvantage, and bold values indicate statistically significant differences after Holm correction.

\begin{table}[htb!]
\centering
\caption{Public text-watermark evidence and within-generation self-detection results.}
\label{tab:watermark-llms}
\scriptsize
\begin{tabular}{ccp{0.55\linewidth}c}
\toprule
Generation & Model & Public evidence for text watermarking on the evaluated endpoint & $\Delta$Acc (pp).\\
\midrule
\multirow{2}{*}{G1}
 & GPT-3.5 Turbo Instruct & Text watermark researched by provider; no deployment confirmed & $+0.18$\\
 & Mixtral 8$\times$22B Instruct & No public model- and endpoint-specific claim identified & $-1.85$\\
\midrule
\multirow{5}{*}{G2}
 & DeepSeek-V3 & No public model- and endpoint-specific claim identified & $\mathbf{-8.15}$\\
 & GLM-4 32B 0414 & No public model- and endpoint-specific claim identified & $-2.06$\\
 & GPT-4o & Text watermark researched by provider; no deployment confirmed & $-0.99$\\
 & Llama 3.3 70B Instruct & No public model- and endpoint-specific claim identified & $\mathbf{+2.96}$\\
 & Qwen2-72B & No public model- and endpoint-specific claim identified & $+0.01$\\
\midrule
\multirow{8}{*}{G3}
 & Claude Opus 4.7 & Provider-level rollout announced; not confirmed for this pre-transition model endpoint & $\mathbf{+0.86}$\\
 & DeepSeek-V4-Pro & No public model- and endpoint-specific claim identified & $+0.05$\\
 & Gemini 3.1 Pro Preview & SynthID-Text documented for Gemini app/web; not confirmed for the evaluated API route & $\mathbf{-5.06}$\\
 & GLM-5.1 & No public model- and endpoint-specific claim identified & $\mathbf{-5.76}$\\
 & GPT-5.5 & Text watermark researched by provider; no deployment confirmed & $+0.1$\\
 & Kimi K2.6 & No public model- and endpoint-specific claim identified & $+0.35$\\
 & MiMo V2.5 Pro & No public model- and endpoint-specific claim identified & $\mathbf{+5.90}$\\
 & Qwen3.6 Max Preview & No public model- and endpoint-specific claim identified & $\mathbf{+1.86}$\\
\bottomrule
\end{tabular}
\end{table}

\end{document}